\documentclass[lettersize,journal]{IEEEtran}
\usepackage{amsmath,amsfonts}
\usepackage{algorithmic}
\usepackage{algorithm}
\usepackage{array}
\usepackage[caption=false,font=normalsize,labelfont=sf,textfont=sf]{subfig}
\usepackage{textcomp}
\usepackage{stfloats}
\usepackage{url}
\usepackage{verbatim}
\usepackage{graphicx}
\usepackage{cite}
\usepackage{multirow}
\usepackage{url}
\usepackage{hyperref}

\usepackage{pifont}
\usepackage{multirow}
\usepackage{hhline, booktabs}
\usepackage{colortbl}
\usepackage{float} 
\usepackage{enumerate}
\usepackage{comment}
\usepackage{tikz}
\definecolor{MyColorTab}{RGB}{230, 230, 230}
\definecolor{ForestGreen}{RGB}{34,139,34}
\definecolor{OrangeRed}{RGB}{236,83,83}
\definecolor{Dangerous}{RGB}{250,208,189}
\definecolor{Best}{RGB}{120,209,115}
\definecolor{SecondBest}{RGB}{206,237,204}

\begin{document}

\title{Rectifying Geometry-Induced Similarity Distortions for \emph{Real-World} Aerial–Ground Person Re-Identification}

\author{
Kailash A. Hambarde,
Hugo Proença,~\IEEEmembership{Senior Member,~IEEE}
\thanks{Manuscript received February XX, 2025; revised XX XX, 2025.}
\thanks{This work was supported by national funds through FCT – Fundação para a Ciência e a Tecnologia, I.P., and, when applicable, co-funded by EU funds under project UID/50008/2025 – Instituto de Telecomunicações.}
\thanks{Kailash A. Hambarde and Hugo Proença are with the Instituto de Telecomunicações and the Department of Computer Science, Universidade da Beira Interior, Portugal.}
\thanks{Corresponding author: Kailash A. Hambarde (e-mail: \href{mailto:kailash.hambarde@ubi.pt}{kailash.hambarde@ubi.pt}).}
}

\markboth{Submitted to IEEE Transactions on Information Forensics and Security}
{Kailash A. Hambarde \MakeLowercase{\textit{et al.}}: Rectifying Geometry-Induced Similarity Distortions for \emph{Real-World} Aerial–Ground Person Re-Identification}

\maketitle

\begin{abstract}
Aerial--ground person re-identification (AG-ReID) is fundamentally challenged by extreme viewpoint and distance discrepancies between aerial and ground cameras, which induce severe geometric distortions and invalidate the assumption of a shared similarity space across views. 
 The existing methods primarily use geometry-aware feature learning or appearance-conditioned prompting, and implicitly assume that the geometry-invariant dot-product similarity used within ''attention'' remains reliable under large viewpoint and distance variations. 
The main hypotheis in this work is that this assumption does not hold: extreme camera geometry and scale variations systematically distort the query--key similarity space, and heavily compromise attention-based matching, even when feature representations are partially aligned.
To address this issue, we introduce the notion of \emph{Geometry-Induced Query--Key Transformation} (GIQT), a lightweight and low-rank module that explicitly rectifies the similarity space by conditioning query-key interactions on camera geometry. 
Rather than modifying the feature content or the attention formulation itself, GIQT adapts the similarity computation to compensate for dominant, geometry-induced anisotropic distortions. 
Building on this local similarity rectification, we consider a geometry-conditioned prompt generation mechanism as an auxiliary component that provides global and view-adaptive representation priors, that are derived directly from camera geometry.
Our experiments were carried out on four aerial-ground person re-identification benchmarks, and show that the proposed framework consistently improves the re-identification  robustness under extreme and even previously unseen geometric perspectives. In pracice, this work gives a step towards operation in high-altitude and large viewing-angle regimes. Also, such contribution is provided at minimal computational overheads, with respect to the state-of-the-art methods. {https://github.com/kailashhambarde/GeoReID.git}
\end{abstract}

\section{Introduction}
\label{sec:intro}

\begin{figure}
    \centering
    \includegraphics[width=\linewidth]{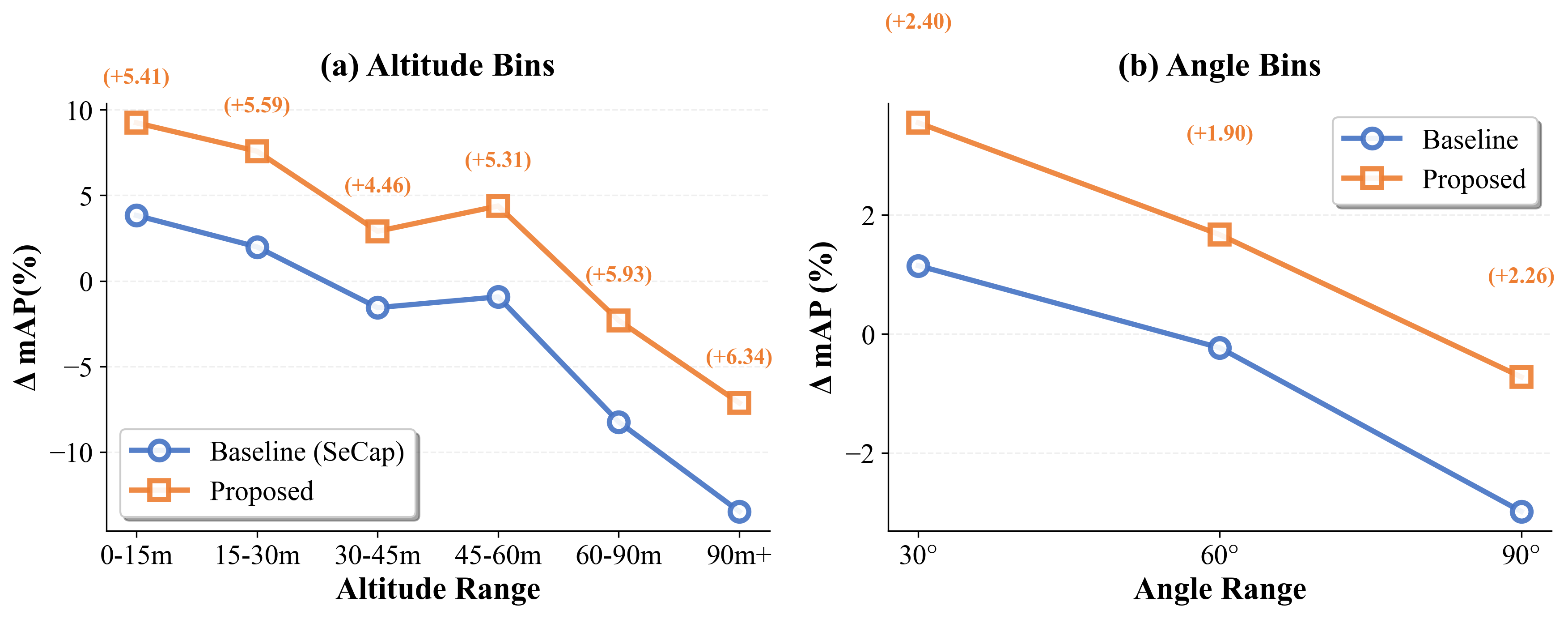}
    \caption{
        Comparison between the performance degradation of the well known SeCap method as baseline and our proposal, across altitude and angle bins.
        The y-axis povide the $\Delta$ mAP = (Bin mAP - Overall SeCap A→G mAP). 
        Values above zero indicate \emph{better-than-overall} performance and values below zero indicate perfomance degradation. Overall, it is evident that the proposed method keeps superior performance in extreme geometric regimes.
    }
    \label{fig:degradation_analysis}
\end{figure}

\IEEEPARstart{P}{erson} re-identification (ReID) is a classical problem of computer vision/machine learning domains, and aims to match identities across non-overlapping cameras~\cite{zhang2023ground, chen2018camera, zheng2016survey}. Here, 
most significant progresses have been achieved in ground-to-ground ReID, using emerging network architectures and training strategies~\cite{li2022pyramidal, yang2022sampling, chen2021person}. 
However, as ground-based camera networks remain limited in spatial coverage, efforts have been concentrated in using unmanned aerial vehicles (UAVs) to broaden surveillance capabilities~\cite{zhang2020person, li2021uav} of such type of systems. 
The combination of aerial and ground cameras gave rise to the aerial-ground person re-identification (AG-ReID) problem, which requires to match identities across drastically different viewpoints and distances.
AG-ReID is substantially more challenging than conventional ReID due to extreme perspective changes: aerial images often exhibit top-down or oblique views, while ground images typically capture frontal or profile appearances.
Such viewpoint discrepancies induce severe geometric distortions, including scale compression, foreshortening, and body-part displacement, yielding large appearance variation and unreliable spatial correspondence across views~\cite{tan2024rle, li2021diverse}.
Beyond appearance variation, extreme camera geometry exposes a fundamental limitation of the existing ReID pipelines: the assumption of a shared similarity space across views.
Modern transformer based ReID models rely on dot-product attention to compute similarity between local features~\cite{dosovitskiy2020image, he2021transreid}.
Even when feature representations are semantically aligned, viewpoint geometry warps spatial correspondence and feature orientation in a highly anisotropic manner~\cite{teng2021viewpoint, zhang2024view}.
As a result, the geometry-invariant dot-product similarity used within attention becomes unreliable: corresponding regions may yield low similarity, while unrelated regions may spuriously align.
Importantly, this does not indicate a limitation of attention or cross-attention mechanisms themselves, but rather a mismatch between extreme camera geometry and the default similarity metric used to compute attention weights.
This behavior has been implicitly observed in prior AG-ReID studies, where performance degrades sharply as geometric disparity increases~\cite{nguyen2023aerial, nguyen2024ag}.
This effect is empirically illustrated in Fig.~\ref{fig:degradation_analysis}, where performance degradation correlates monotonically with altitude and viewing-angle differences.
The increasing performance gap under extreme geometry indicates that camera geometry introduces structured distortion in similarity computation rather than random noise.
The existing AG-ReID methods mitigate cross-view discrepancies using attribute supervision~\cite{nguyen2023aerial}, view-disentangled representations~\cite{zhang2024view}, or prompt-based feature adaptation.
While effective in moderate settings, these approaches primarily rely on implicit learning of geometric relationships and typically ignore explicit camera geometry during similarity computation, limiting robustness under extreme or unseen viewpoints.

Motivated by these observations, we argue that robust AG-ReID requires explicitly accounting for geometry-induced distortion in cross-view similarity computation, rather than further modifying feature representations.
To this end, we propose a geometry-conditioned similarity alignment framework that incorporates camera geometry into both global representation adaptation and local similarity computation.

The main contributions of this work are four-fold:

\begin{itemize}
    \item We identify similarity-space distortion as a dominant failure mode in aerial--ground person re-identification under extreme camera geometry, showing that the geometry-invariant similarity assumption underlying attention becomes unreliable even when features are reasonably aligned.

    \item We propose a geometry-conditioned similarity alignment framework that explicitly adapts cross-view similarity computation rather than relying solely on feature adaptation.

    \item We introduce the Geometry-Induced Query--Key Transformation (GIQT), a lightweight and model-agnostic module that reshapes the attention similarity space via a geometry-conditioned low-rank transformation.

    \item We provide empircal evidence about a substantially stronger robustness and generalization under extreme and even unseen geometric conditions with respect to the state-of-the-art. Such improvements are observed across multiple AG-ReID benchmarks.
\end{itemize}


\section{Related Work}
\label{rel_work}

\subsection{Aerial--Ground Person Re-Identification}

Aerial-ground ReID extends conventional ReID to scenarios involving UAV-mounted and ground-based cameras, introducing extreme viewpoint, scale, and orientation discrepancies that fundamentally violate assumptions made in ground-to-ground settings~\cite{nguyen2023aerial, hambarde2024image}. 
Unlike conventional ReID, where viewpoint variations are relatively limited, AG-ReID must cope with severe geometric distortions caused by altitude changes, oblique viewing angles, and drastic perspective shifts.
Early research in AG-ReID primarily focused on benchmark construction to enable systematic evaluation.
Nguyen et al. introduced the AG-ReID.v1 dataset~\cite{nguyen2023aerial} and its extension AG-ReID.v2~\cite{nguyen2024ag}, highlighting the difficulty of matching identities across aerial and ground views. 
Zhang et al. proposed the large-scale synthetic CARGO benchmark to explicitly stress-test cross-view appearance changes under extreme geometry~\cite{zhang2024view}.
Subsequent video-based benchmarks, such as G2A-VReID and AG-VPReID, incorporated temporal cues while retaining the same fundamental geometric challenges~\cite{zhang2024cross, nguyen2025person}. 
These datasets demonstrate that appearance-driven representations are inherently fragile under extreme viewpoint geometry.

\subsection{Implicit Geometry Modelling in AG-ReID}

Most existing AG-ReID methods address viewpoint discrepancies through implicit feature learning.
Multi-stream architectures separate aerial and ground representations and rely on auxiliary supervision, such as identity attributes, to encourage view-invariant encoding~\cite{nguyen2024ag}. 
Transformer-based approaches, including the View-Decoupled Transformer (VDT), attempt to disentangle identity related and viewpoint related factors using view tokens and orthogonality constraints~\cite{zhang2024view}.
Prompt based methods, such as SeCap, further adapt representations through appearance conditioned prompts or dynamic token selection~\cite{wang2025secap}.
While effective in practice, these approaches learn geometric relationships implicitly from data. 
Viewpoint geometry such as camera altitude, viewing angle, or sensor configuration, is not explicitly modelled, limiting interpretability and control over how geometric distortions influence feature alignment. 
As a result, such methods often struggle to generalise under unseen or extreme viewpoint configurations.
\subsection{Geometry Aware and View Invariant Learning}

Several studies have explored mechanisms related to geometric robustness in cross view recognition. 
Adversarial domain adaptation has been employed to reduce distribution gaps between aerial and ground features~\cite{khalid2025bridging}, while view aware attention mechanisms attempt to capture view dependent transformations implicitly~\cite{zhang2024view}. 
Diffusion based synthesis augments training data with generated cross view samples~\cite{wang2025sdreid}, and disentangled representation learning separates identity from view specific factors~\cite{nguyen2024ag}. 
Rotation robustness has also been investigated through equivariant architectures and rotation-aware tokens~\cite{zhang2025dari, wang2024rotation, chen2025towards}.
However, these approaches primarily rely on data augmentation or implicit representation learning and do not explicitly incorporate known geometric priors. 
Crucially, readily available metadata in aerial sensing, such as altitude, viewing angle, and camera identity, are typically ignored during both representation learning and attention-based similarity computation.

\subsection{Geometric Alignment and Multi-Modal Cues}

Geometric and semantic alignment methods aim to reduce cross view mismatch by establishing correspondences across views. 
GSAlign leverages coupled geometric semantic alignment networks to learn dense correspondences~\cite{li2025gsalign}, while part based and keypoint driven approaches rely on pose estimation to guide alignment~\cite{qiu2024salient, wang2024dynamic}. 
However, keypoint detection and part localization are unreliable in low resolution aerial imagery and under severe occlusion, limiting their effectiveness in AG-ReID scenarios.
Multi modal approaches introduce textual or language-based supervision to enhance view invariance~\cite{wang2025aea, zhang2025latex, li2025mmreid}. 
While promising, such methods increase system complexity and often depend on additional annotations or pretrained language models, making them less suitable for lightweight UAV-based deployment.

Despite significant progress, existing AG-ReID approaches share several limitations. 
First, geometric relationships are typically modeled implicitly, limiting interpretability and controllability. 
Second, robustness degrades substantially under extreme altitude and viewing-angle variations. 
Third, many methods introduce considerable computational overhead, hindering real-time UAV deployment. 
Finally, generalization across datasets with different camera geometries remains limited.
These limitations motivate the need for explicitly geometry aware AG-ReID frameworks that directly incorporate aerial sensing priors into both global representation adaptation and local similarity computation, rather than relying solely on implicit feature learning.


\begin{figure*}[ht]
    \centering
    \includegraphics[width=\textwidth]{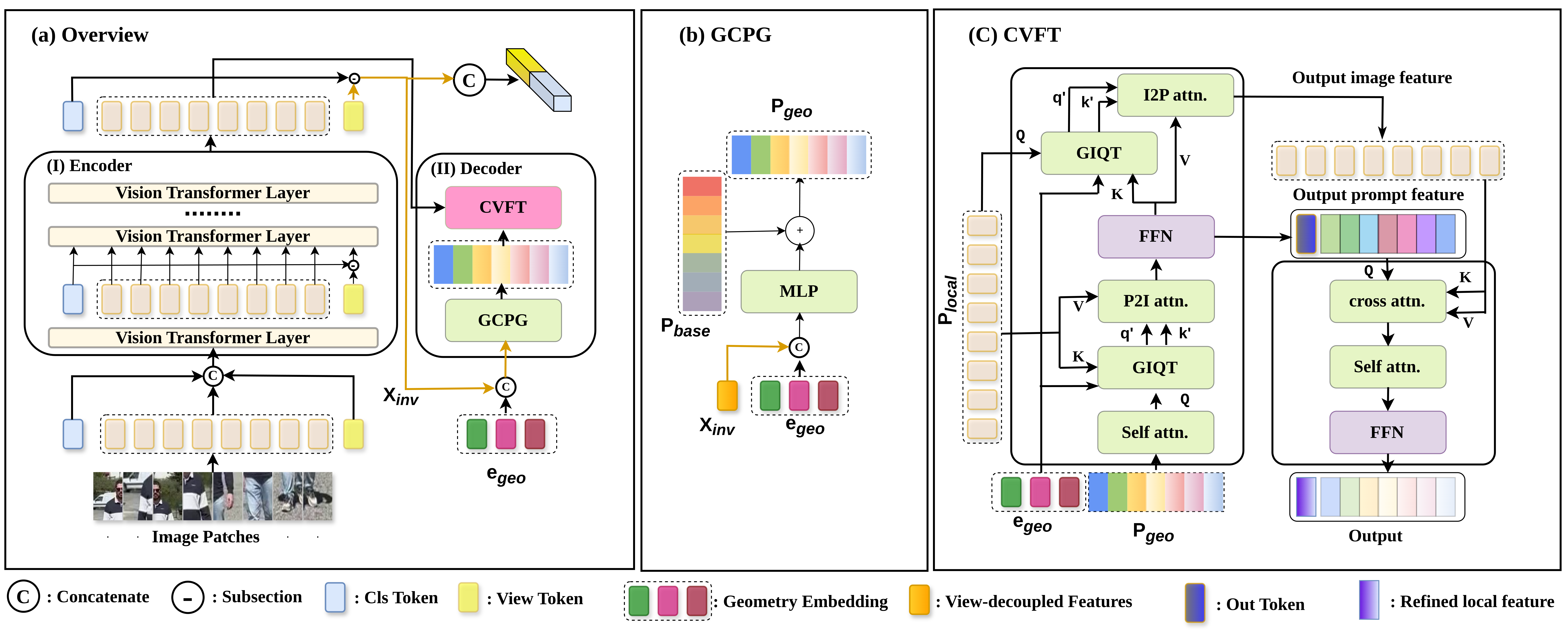}
    \caption{
    (a) The architecture of the proposed geometry conditioned similarity alignment framework. The encoder (VDT) extracts visual features and decouples view-related and view-invariant components; the decoder comprises GCPG and CVFT. The final ReID representation is $\mathbf{Out} = [X_{\mathrm{inv}}, X_{\mathrm{ref}}]$ (view-invariant global descriptor plus refined local features from CVFT). (b) Geometry Conditioned Prompt Generation (GCPG): inputs are the view-invariant descriptor $X_{\mathrm{inv}}$ and geometry embedding $\mathbf{e}_{\mathrm{geo}}$; an MLP $f_{\mathrm{geo}}$ yields geometry-conditioned prompts $P_{\mathrm{geo}} = P_{\mathrm{base}} + \alpha \cdot f_{\mathrm{geo}}(X_{\mathrm{inv}}, \mathbf{e}_{\mathrm{geo}})$. (c) Cross-View Feature Transformation (CVFT, LFRM): two-way attention with GIQT decodes discriminative local features from $X_{\mathrm{local}}$ using $P_{\mathrm{geo}}$ and $\mathbf{e}_{\mathrm{geo}}$.
    }
    \label{fig:approach}
\end{figure*}

\begin{figure}[ht]
    \centering
    \includegraphics[width=\linewidth, height=4cm]{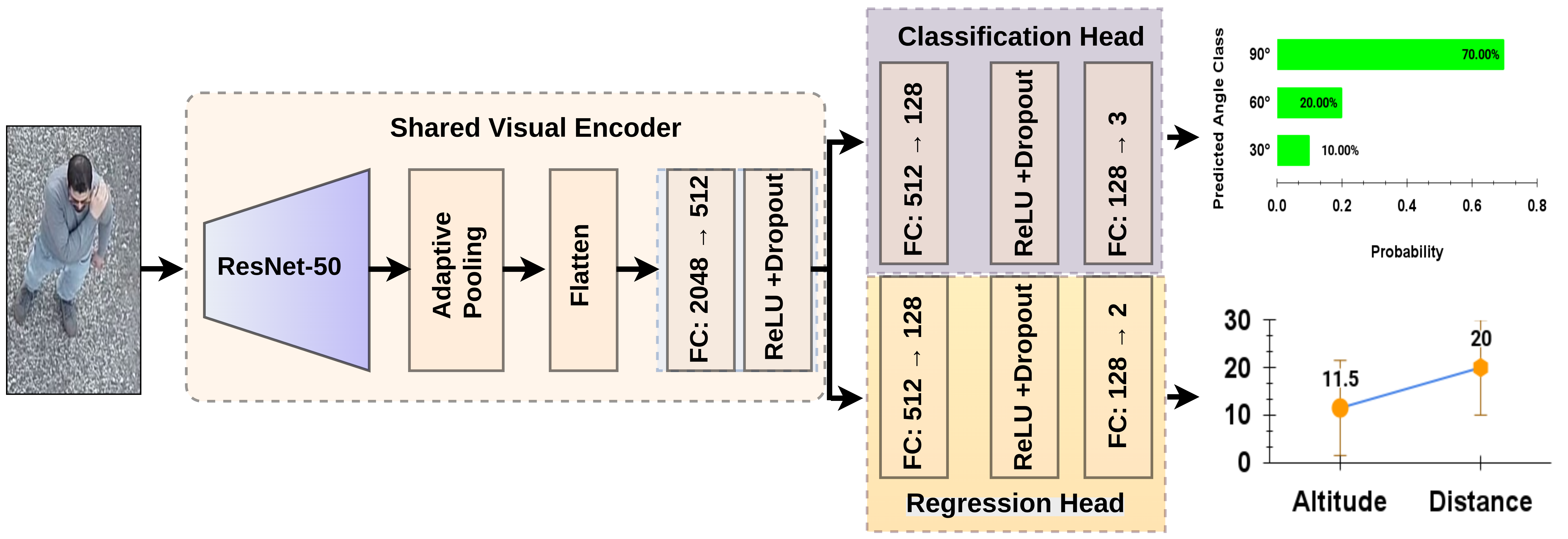}
    \caption{Vision-only multi-task geometry prediction. A shared ResNet-50 encoder extracts visual features from RGB images, followed by task-specific heads for altitude and distance regression and viewing-angle classification.}
    \label{fig:metadata_predictor}
\end{figure}

\section{Method}
\label{method}

\subsection{Overview}

The overall framework of the proposed geometry conditioned similarity alignment framework, as illustrated in Fig.~\ref{fig:approach} (a), adopts an encoder-decoder transformer architecture. The encoder is the View Decoupling Transformer (VDT) \cite{zhang2024view}. In contrast to the conventional ViT \cite{liu2021swin}, our approach incorporates the \textbf{View} token and performs hierarchical decoupling of the \textbf{Cls} token at each layer, effectively segregating view-related and view-invariant features within the \textbf{Cls} token, while extracting local features from the input \cite{wang2025secap}.
The decoder comprises the Geometry Conditioned Prompt Generation module (\textbf{GCPG}) and the Cross-View Feature Transformation module (\textbf{CVFT}). The \textbf{GCPG} adaptively generates prompts for different viewpoints based on the view invariant descriptor $X_{\mathrm{inv}}$ and the camera geometry embedding $\mathbf{e}_{\mathrm{geo}}$. The \textbf{CVFT} utilizes the geometry conditioned prompts $P_{\mathrm{geo}}$ and applies the Geometry Induced Query Key Transformation (\textbf{GIQT}) inside cross attention to decode geometry aligned local features. 
The final ReID representation is $\mathbf{Out} = [X_{\mathrm{inv}}, X_{\mathrm{ref}}]$, i.e., view-invariant global descriptor plus refined local features; the encoder outputs $\mathbf{Cls}$, $\mathbf{View}$, and $X_{\mathrm{local}}$ are used only internally as in Eq.~(1). The overall framework can be described as follows:
\begin{equation}
    \begin{aligned}
    [\mathbf{Cls}, \mathbf{View}, X_{\mathrm{local}}] &= \mathrm{VDT}([\mathbf{CLS}, \mathbf{View}, \mathrm{tokenization}(\mathbf{X})]), \\
    X_{\mathrm{inv}} &= \mathbf{Cls} - \mathbf{View}, \\
    \mathbf{e}_{\mathrm{geo}} &= [\mathbf{e}_{\mathrm{cam}};\, \mathbf{e}_{\mathrm{alt}};\, \mathbf{e}_{\mathrm{angle}}], \\
    P_{\mathrm{geo}} &= \mathcal{G}_{\mathrm{prompt}}(X_{\mathrm{inv}}, \mathbf{e}_{\mathrm{geo}}), \\
    X_{\mathrm{ref}} &= \mathcal{G}_{\mathrm{align}}(X_{\mathrm{local}}, P_{\mathrm{geo}}, \mathbf{e}_{\mathrm{geo}}), \\
    \mathbf{Out} &= [X_{\mathrm{inv}}, X_{\mathrm{ref}}],
    \end{aligned}
\end{equation}
where VDT is the View Decoupling Transformer; $\mathbf{Cls}$ and $\mathbf{View}$ are the class token and view token output by the encoder; $\mathrm{tokenization}(\mathbf{X})$ denotes the process of converting the input image $\mathbf{X}$ into patch tokens; $X_{\mathrm{local}} \in \mathbb{R}^{B \times N \times d}$ denotes the local patch features ($B$ batch size, $N$ number of patches, $d$ feature dimension); $X_{\mathrm{inv}}$ is the view-invariant global descriptor; $\mathbf{e}_{\mathrm{geo}}$ is the geometry embedding encoding altitude, viewing angle, and camera identity; $P_{\mathrm{geo}}$ denotes the geometry conditioned prompts; $\mathcal{G}_{\mathrm{prompt}}$ is the prompt generation module (GCPG); $\mathcal{G}_{\mathrm{align}}$ is the local refinement module (CVFT) that uses GIQT; $X_{\mathrm{ref}}$ is the refined local features; and $\mathbf{Out}$ is the final representation used for ReID.


\subsection{Geometry Metadata Acquisition}

Our framework relies on camera geometry cues: altitude, viewing angle, and camera identity. When such metadata is provided by the dataset, we directly encode the ground truth geometry. When metadata is unavailable or incomplete, we train a vision only multi task geometry prediction network to estimate these cues from RGB images. 
As shown in Fig.~\ref{fig:metadata_predictor}, a shared ResNet-50 encoder extracts visual features, followed by task-specific heads for altitude and distance regression and viewing-angle classification. The predicted geometry is used consistently during training and inference, enabling geometry-conditioned alignment without external sensors.


\subsection{Geometry Conditioned Prompt Generation (GCPG)}

Global cross view alignment requires adapting high level representation priors to camera geometry. To this end, we introduce a geometry conditioned prompt generation module that maps the view invariant descriptor and geometry embedding to global prompt offsets. As illustrated in Fig.~\ref{fig:approach} (b), the module first encodes geometry cues into a single embedding vector. Geometry cues are encoded as:
\begin{equation}
    \mathbf{e}_{\mathrm{geo}} = [\mathbf{e}_{\mathrm{cam}};\, \mathbf{e}_{\mathrm{alt}};\, \mathbf{e}_{\mathrm{angle}}] \in \mathbb{R}^{d_{\mathrm{geo}}},
\end{equation}
where $\mathbf{e}_{\mathrm{cam}} \in \mathbb{R}^{d_{\mathrm{cam}}}$ is the camera identity embedding from a learnable table indexed by camera ID $c$; $\mathbf{e}_{\mathrm{alt}} \in \mathbb{R}^{d_{\mathrm{alt}}}$ is the altitude embedding obtained by discretizing altitude $h$ into $N_{\mathrm{alt}}$ bins and indexing a learnable table $\mathbf{E}_{\mathrm{alt}}$; $\mathbf{e}_{\mathrm{angle}} \in \mathbb{R}^{d_{\mathrm{angle}}}$ is the viewing angle embedding from discretizing angle $\theta$ into $N_{\mathrm{angle}}$ bins and indexing $\mathbf{E}_{\mathrm{angle}}$; and $d_{\mathrm{geo}} = d_{\mathrm{cam}} + d_{\mathrm{alt}} + d_{\mathrm{angle}}$.
Given the view invariant descriptor $X_{\mathrm{inv}}$ and geometry embedding $\mathbf{e}_{\mathrm{geo}}$, geometry-conditioned prompts are generated as:
\begin{equation}
    P_{\mathrm{geo}} = P_{\mathrm{base}} + \alpha \cdot f_{\mathrm{geo}}(X_{\mathrm{inv}}, \mathbf{e}_{\mathrm{geo}}),
\end{equation}
where $P_{\mathrm{base}} \in \mathbb{R}^{L \times d}$ denotes learnable base prompts ($L$ prompt length, $d$ feature dimension); $\alpha$ is a learnable scaling factor; and $f_{\mathrm{geo}}(\cdot)$ is a lightweight geometry conditioned mapping network. The residual formulation preserves identity discriminative semantics while injecting geometry as a structured bias. Geometry conditioned prompts act as global priors that guide the decoder toward geometry consistent cues under varying viewpoints.


\subsection{Geometry Induced Query Key Transformation (GIQT)}

While geometry conditioned prompts provide coarse global adaptation, extreme viewpoint changes introduce local geometric distortions that challenge the geometry invariant similarity assumption used in standard attention. 
The proposed Geometry Induced Query Key Transformation addresses this by adapting the similarity metric used to compute attention weights. Empirical analysis of cross view feature covariance reveals pronounced spectral decay, indicating that viewpoint induced distortions are highly anisotropic; a small number of dominant components capture most geometry induced variation, motivating a low rank correction strategy.
Before computing attention, geometry conditioned transformations are applied to queries and keys and values are kept unchanged to preserve feature content:
\begin{equation}
    Q' = T_Q(\mathbf{e}_{\mathrm{geo}}) Q, \qquad K' = T_K(\mathbf{e}_{\mathrm{geo}}) K.
\end{equation}
Where $Q \in \mathbb{R}^{L \times d_k}$ and $K \in \mathbb{R}^{N \times d_k}$ are the query and key matrices per head ($d_k = d/H$ is the head dimension, $H$ the number of heads); $T_Q(\mathbf{e}_{\mathrm{geo}})$ and $T_K(\mathbf{e}_{\mathrm{geo}}) \in \mathbb{R}^{d_k \times d_k}$ are learned transformation matrices conditioned on $\mathbf{e}_{\mathrm{geo}}$; and $Q'$, $K'$ are the transformed query and key. In implementation, $T_Q$ and $T_K$ are applied \emph{per head} to the head dimension vectors. Attention is then computed as $\mathrm{Attention}(Q', K', V) = \mathrm{softmax}(Q' (K')^\top / \sqrt{d_k}) V$, with $V$ unchanged. To avoid over-parameterization and stabilize training, we adopt a low-rank residual formulation:
\begin{equation}
    T(\mathbf{e}_{\mathrm{geo}}) = I + U(\mathbf{e}_{\mathrm{geo}}) V(\mathbf{e}_{\mathrm{geo}})^\top.
\end{equation}
Where $I$ is the identity matrix; $U$ and $V$ are predicted from $\mathbf{e}_{\mathrm{geo}}$ via a lightweight geometry encoder and per head linear predictors, with $U, V \in \mathbb{R}^{d_k \times r}$ and rank $r \ll d_k$. This formulation targets dominant geometry sensitive directions. The transformation can be applied efficiently without materializing $d_k \times d_k$ matrices: for a vector $\mathbf{q}$, $\mathbf{q}' = \mathbf{q} + U(V^\top \mathbf{q})$. Optionally, per head residual gating blends transformed and original features: $\mathbf{q}_{\mathrm{out}} = \alpha^{(h)} \mathbf{q}' + (1 - \alpha^{(h)}) \mathbf{q}$, where $\alpha^{(h)} = \sigma(\beta^{(h)})$ is a learnable scalar per head. By adapting the comparison space rather than feature content, GIQT suppresses unstable viewpoint-dependent similarity directions and emphasizes geometry consistent cues.


\subsection{Geometry Aligned Local Feature Refinement (CVFT)}

The Cross View Feature Transformation (CVFT), is the transformer based decoder, as illustrated in Fig.~\ref{fig:approach} (c). It extracts geometry aligned discriminative features from the local features $X_{\mathrm{local}}$ using the geometry conditioned prompts $P_{\mathrm{geo}}$ from the GCPG and the GIQT inside cross attention. The CVFT consists of the two way attention module and the feature fusion module. The two way attention module employs self-attention (SA) and cross-attention (CA) in both prompt-to-image and image-to-prompt directions; all cross attention operations use GIQT transformed queries and keys. The two way attention module can be described as follows:
\begin{equation}
    \begin{aligned}
    F_{\mathrm{P}} &= \mathrm{FFN}(\mathrm{CA}(\mathrm{SA}(P_{\mathrm{geo}}), X_{\mathrm{local}}, X_{\mathrm{local}})) + P_{\mathrm{geo}}, \\
    F_{\mathrm{I}} &= \mathrm{CA}(X_{\mathrm{local}}, F_{\mathrm{P}}, F_{\mathrm{P}}) + X_{\mathrm{local}},
    \end{aligned}
\end{equation}
where $F_{\mathrm{P}}$ represents the prompt features output by the two-way attention module; $F_{\mathrm{I}}$ denotes the image (local) features output by the two-way attention module; $\mathrm{FFN}$ stands for the Feed-Forward Network; $\mathrm{CA}$ is the cross-attention mechanism (with GIQT applied to queries and keys before the attention score computation); $\mathrm{SA}$ is the self-attention mechanism; $X_{\mathrm{local}}$ signifies the local patch features from the encoder; and $P_{\mathrm{geo}}$ refers to the geometry-conditioned prompts. The feature fusion module employs cross-attention and self-attention to integrate the output token with the prompt and image features, decoding the final local representation. The feature fusion module can be described as follows:
\begin{equation}
    [\mathbf{Out}, \_] = \mathrm{FFN}(\mathrm{SA}(\mathrm{CA}([\mathbf{Out}_{\mathrm{token}}, F_{\mathrm{P}}], F_{\mathrm{I}}, F_{\mathrm{I}}))),
\end{equation}
where $\mathbf{Out}$ is the final local descriptor used for ReID; $\mathbf{Out}_{\mathrm{token}}$ is a learnable output token; $F_{\mathrm{P}}$ is the prompt features; and $F_{\mathrm{I}}$ is the refined image features. The resulting refined features are explicitly aligned with respect to camera geometry, yielding robust local representations under extreme cross-view variations.


\subsection{Optimization}
\label{sec:optimization}

In addition to the ID classification loss and the Triplet loss commonly used in ReID tasks, our loss functions are further enhanced with the view classification loss, the orthogonality loss, and the geometry conditioned prompt regularization \cite{wang2025secap}.

\textbf{View Classification Loss:} To achieve decoupling of view-related features, we utilize a view classifier and employ the view classification loss for constraint. The loss function can be formalized as follows:
\begin{equation}
    \mathcal{L}_{\mathrm{view}} = -\sum_{i=1}^{N} y_i \log(p_i),
\end{equation}
where $y_i$ is the ground-truth view label for the $i$-th sample; $p_i$ is the predicted probability of the $i$-th sample belonging to the correct view class; and $N$ is the total number of samples in the batch.

\textbf{Orthogonality Loss:} To ensure thorough view decoupling, we introduce the orthogonality loss. The expression is as follows:
\begin{equation}
    \mathcal{L}_{\mathrm{orth}} = \sum_{i=1}^{d} \bigl| \langle \mathbf{inv}_i, \mathbf{v}_i \rangle \bigr|,
\end{equation}
where $|\langle \cdot, \cdot \rangle|$ denotes the absolute value of the dot product; $d$ represents the feature dimensionality; $\mathbf{inv}_i$ and $\mathbf{v}_i$ denote the $i$-th component of the view-invariant and view-related feature vectors, and the sum runs over the feature dimension $d$.

\textbf{Geometry-Conditioned Prompt Regularization:} To stabilize geometry-conditioned prompt adaptation, an $\ell_2$ regularization term is applied to the prompt offset:
\begin{equation}
    \mathcal{L}_{\mathrm{geo}} = \bigl\| f_{\mathrm{geo}}(X_{\mathrm{inv}}, \mathbf{e}_{\mathrm{geo}}) \bigr\|_2^2,
\end{equation}
where $f_{\mathrm{geo}}$ is the geometry-conditioned mapping network and the offset is the generated prompt component before adding to $P_{\mathrm{base}}$.

\textbf{Overall Loss:} We apply the ID classification loss and the Triplet loss to both global and local features. The overall optimization objective can be summarized as follows:
\begin{equation}
    \begin{aligned}
    \mathcal{L} &= \alpha \bigl( \mathcal{L}_{\mathrm{ID}}^{\mathrm{global}} + \mathcal{L}_{\mathrm{Tri}}^{\mathrm{global}} \bigr) + \beta \bigl( \mathcal{L}_{\mathrm{ID}}^{\mathrm{local}} + \mathcal{L}_{\mathrm{Tri}}^{\mathrm{local}} \bigr) \\
                &\quad + \lambda \bigl( \mathcal{L}_{\mathrm{view}} + \mathcal{L}_{\mathrm{orth}} \bigr) + \gamma \mathcal{L}_{\mathrm{geo}},
    \end{aligned}
\end{equation}
where $\alpha$ and $\beta$ are hyperparameters that balance the optimization objectives of global and local features; $\lambda$ balances view disentanglement (view classification and orthogonality) with identity learning; and $\gamma$ controls the geometry-conditioned prompt regularization. We set $(\alpha, \beta, \lambda, \gamma) = (1.0, 1.0, 0.5, 0.1)$ in all experiments.




\section{Experiments}
\label{experiment}

\subsection{Datasets and Evaluation Protocols}

\textbf{Datasets.}
We evaluate our approach on four aerial-ground person ReID benchmarks:
AG-ReIDv1~\cite{nguyen2023aerial},
AG-ReIDv2~\cite{nguyen2024ag},
CARGO~\cite{zhang2024view}, and
DetReIDX~\cite{hambarde2025detreidx}.
AG-ReIDv1 and AG-ReIDv2 provide ground truth camera geometry annotations, including drone altitude and camera identity.
DetReIDX provides drone altitude, viewing angles, and camera identity.
In contrast, CARGO only provides camera identity and does not include explicit geometric metadata such as altitude or viewing angle.
The CARGO dataset contains 108,563 images of 5,000 identities captured by eight ground cameras and five aerial cameras.
AG-ReIDv1 consists of 21,983 images of 388 identities collected using one aerial camera and one ground camera, with aerial views captured at altitudes ranging from 15 to 45 meters.
AG-ReIDv2 extends AG-ReIDv1 by introducing additional viewpoints, more identities, and more diverse camera configurations.
DetReIDX contains 553 identities and includes one training split and three evaluation protocols:
A$\rightarrow$A (105,478 images),
A$\rightarrow$G (72,033 images), and
G$\rightarrow$A (72,033 images).
The dataset is collected across six international university campuses and features substantial variation in drone altitude and viewing angle.
For CARGO, we use the proposed vision-only geometry predictor to generate altitude, horizontal distance, and viewing angle.
For AG-ReIDv1 and AG-ReIDv2, the geometry predictor is used only to estimate viewing angles when ground-truth angle annotations are unavailable.

\textbf{Evaluation Protocols.}
Following common practice in aerial-ground person ReIDs we evaluate all methods using the
Cumulative Matching Characteristic (CMC) at Rank-1~\cite{moon2001computational} and mean Average Precision (mAP)~\cite{zheng2015scalable}.
For the CARGO dataset, we follow the benchmark defined protocols and report results under four settings:
\emph{ALL}, which evaluates performance across all query--gallery pairs;
\emph{G$\leftrightarrow$G}, which evaluates ground-to-ground matching;
\emph{A$\leftrightarrow$A}, which evaluates aerial-to-aerial matching; and
\emph{A$\leftrightarrow$G}, which evaluates the most challenging aerial-to-ground matching scenario.
For AG-ReIDv1, we adopt the standard bidirectional cross view evaluation protocol between aerial and ground cameras,
namely \emph{A$\rightarrow$G} and \emph{G$\rightarrow$A}.
AG-ReIDv2 extends AG-ReIDv1 by introducing additional camera viewpoints.
In particular, the wearable view (\emph{W}) represents a low viewpoint close to human eye level,
while the ground view (\emph{G}) corresponds to fixed CCTV-style cameras.
Based on these views, AG-ReIDv2 defines four cross view protocols:
\emph{A$\rightarrow$G}, \emph{G$\rightarrow$A}, \emph{A$\rightarrow$W}, and \emph{W$\rightarrow$A}.
For DetReIDX, we follow the official evaluation protocol and report results under three settings:
\emph{A$\rightarrow$A}, \emph{A$\rightarrow$G}, and \emph{G$\rightarrow$A}.
These protocols involve significant variation in drone altitude, viewing angle, and image resolution, making cross-view retrieval particularly challenging.
In all cases, \emph{A} denotes aerial views, \emph{G} denotes ground views, and \emph{W} denotes wearable views.
The arrow indicates the retrieval direction from query to gallery, while the double arrow indicates bidirectional evaluation.
All methods are evaluated using identical train test splits and metrics to ensure fair and reproducible comparison.

\subsection{Implementation Details}

Our proposed method is implemented using PyTorch and the FastReID framework.
All experiments are conducted on NVIDIA A40 GPUs. 
We adopt a pre-trained Vision Transformer Base (ViT-B/16) as the backbone network, initialized with ImageNet pre-trained weights.
All input images are resized to $256 \times 128 \times 3$ pixels. 
For data augmentation, we apply random horizontal flipping, padding, and random erasing \cite{zhong2020random} with a probability of 0.5.
The model is trained using a mini-batch size of 128 (or 192 for larger datasets), where each batch consists of 32 identities with 4 instances per identity (or 8 instances for datasets with more samples per identity). 
The model is optimized using SGD \cite{bottou2010large} with a base learning rate of 0.008 for 120 epochs on the dataset size.
We employ a cosine annealing learning rate schedule with a minimum learning rate of $1.6 \times 10^{-6}$ and a warmup period of 2000 iterations.
The weight decay is set to 0.0001, and gradient clipping is enabled for training stability.
Mixed precision training (Automatic Mixed Precision) is utilized to accelerate training and reduce memory consumption.

\subsection{Comparison with State-of-the-art Methods}

We evaluate the proposed GeoReID framework on four aerial-ground person ReID benchmarks, namely AG-ReIDv1, AG-ReIDv2, CARGO, and DETReIDX, and compare it with a wide range of state-of-the-art methods.
\begin{table}[t]
\centering
\scriptsize
\renewcommand\tabcolsep{4pt}
\renewcommand\arraystretch{1.1}

\caption{\textbf{Performance comparison on AG-ReID under two cross-view protocols.}
All methods are evaluated under identical training and testing settings.
Methods not originally designed for AG-ReID are adapted following their official implementations.
Geometry information is provided only to methods that explicitly support geometry conditioning.
Best results are highlighted as \colorbox{Best}{\textbf{bold}}, and second-best results are highlighted as
\colorbox{SecondBest}{\underline{underlined}}.}

\label{Tbl:agreid_main}

\begin{tabular}{c|cc|cc}
\hline
\scriptsize
\multirow{2}{*}{\textbf{Method}}
    & \multicolumn{2}{c|}{Protocol 1: A$\leftrightarrow$G}
    & \multicolumn{2}{c}{Protocol 2: G$\leftrightarrow$A} \\
\cline{2-5}
    & Rank1 & mAP & Rank1 & mAP \\
\hline
SBS~\cite{he2023fastreid}          & 73.54 & 59.77 & 73.70 & 62.27 \\
BoT~\cite{luo2019bag}              & 70.01 & 55.47 & 71.20 & 58.83 \\
OsNet~\cite{zhou2021learning}      & 72.59 & 58.32 & 74.22 & 60.99 \\
VV~\cite{kuma2019vehicle} & 77.22 & 67.23 & 79.73 & 69.83 \\
Explain  \cite{nguyen2023ag} & 81.28 & 72.38 & 82.64 & 73.35 \\
ViT~\cite{dosovitskiy2020image}    & 81.47 & 72.61 & 82.85 & 73.39 \\

TrasReID  \cite{he2021transreid} &  81.80 & 73.10 & 83.40 & 74.60\\
PFD \cite{wang2022pose}  & 82.30 & 73.60 & 82.50 & 73.90\\
PHA \cite{zhang2023pha} & 79.30 & 71.30 & 81.10 & 72.10\\
FusionReID \cite{wang2025unity} & 80.40 & 71.40 & 82.40 & 74.20\\
PCL-CLIP \cite{li2023prototypical} & 82.16 & 73.11 & 86.90 & 76.28\\
DTST \cite{wang2024dynamic} & 83.48 & 74.51 & 84.72 & 76.05\\
VDT~\cite{zhang2024view} & 82.91 & 74.44 & 86.59 & 78.57 \\
CLIP-ReID~\cite{li2023clip} & 72.61 & 62.09 & 74.12 & 64.19 \\
AG-ReID~\cite{nguyen2023aerial} & 82.91 & 72.38 & 82.85 & 73.35 \\
SeCap~\cite{wang2025secap} & 84.03 & 76.16 & 87.01 & 78.34 \\
LATex \cite{zhang2025latex} & 85.26 & 77.67 & \cellcolor{SecondBest} \underline{89.40} & 81.15 \\
SD-ReID \cite{wang2025sdreid} & 85.16 & 75.40 & 77.02 & \cellcolor{SecondBest} \underline{85.57}\\
GSAlign~\cite{li2025gsalign}  & \cellcolor{SecondBest} \underline{86.74} & \cellcolor{Best}\textbf{84.00} & 87.94 & \cellcolor{Best}\textbf{87.17} \\
\hline
\textbf{GeoReID} & \cellcolor{Best}\textbf{87.02} & \cellcolor{SecondBest} \underline{79.46} & \cellcolor{Best}\textbf{90.64} &  83.55 \\
\hline
\end{tabular}
\end{table}

Quantitative results on AG-ReID are reported in Table~\ref{Tbl:agreid_main}.
GeoReID achieves the highest Rank-1 accuracy under both A$\leftrightarrow$G and G$\leftrightarrow$A evaluation protocols.
Under the A$\leftrightarrow$G setting, GeoReID attains 87.02\% Rank-1 accuracy and 79.46\% mAP, improving upon the strongest prior Rank-1 result while maintaining competitive retrieval quality.
Under the G$\leftrightarrow$A setting, GeoReID further achieves 90.64\% Rank-1 accuracy and 83.55\% mAP, surpassing previous methods in Rank-1 performance.
These results indicate that geometry conditioned similarity alignment consistently improves cross view matching accuracy.
\begin{table}[t]
\centering
\scriptsize
\fontsize{7.5pt}{10pt}\selectfont
\renewcommand\tabcolsep{1.2pt}
\renewcommand\arraystretch{1.12}
\caption{\textbf{Performance comparison on the AG-ReIDv2 dataset across four protocols.}
Best results are highlighted as \protect\colorbox{Best}{\textbf{bold}}, and second-best results are highlighted as
\protect\colorbox{SecondBest}{\underline{underlined}}.}
\label{Tbl:agreidv2_main}
\begin{tabular}{l|cc|cc|cc|cc}
\hline
\multirow{2}{*}{\textbf{Model}} 
    & \multicolumn{2}{c|}{\textbf{A→G}} 
    & \multicolumn{2}{c|}{\textbf{G→A}} 
    & \multicolumn{2}{c|}{\textbf{A→W}} 
    & \multicolumn{2}{c}{\textbf{W→A}} \\
\cline{2-9}
    & Rank1 & mAP & Rank1 & mAP & Rank1 & mAP & Rank1 & mAP \\
\hline
Swin \cite{liu2021swin}         & 68.76 & 57.66 & 68.80 & 57.70 & 68.49 & 56.15 & 64.40 & 53.90 \\
HRNet-18 \cite{wang2020deep}     & 75.21 & 65.07 & 76.25 & 66.16 & 76.26 & 66.17 & 76.25 & 66.17 \\
SwinV2  \cite{liu2022swin}      & 76.44 & 66.09 & 77.11 & 62.14 & 80.08 & 69.09 & 74.53 & 65.61 \\
MGN (R50) \cite{wang2018learning}     & 82.09 & 70.17 & 84.21 & 72.41 & 88.14 & 78.66 & 84.06 & 73.73 \\
BoT (R50) \cite{luo2019bag}     & 80.73 & 71.49 & 79.46 & 69.67 & 86.06 & 75.98 & 82.69 & 72.41 \\
SBS (R50) \cite{he2023fastreid}    & 81.43 & 72.19 & 80.15 & 70.37 & 86.66 & 76.68 & 83.29 & 73.11 \\
ViT \cite{dosovitskiy2020image} & 85.40 & 77.03 & 84.65 & 75.90 & 89.77 & 80.48 & 84.27 & 76.59\\
TransReID \cite{he2021transreid} & 88.00 & 81.40 & 87.60 & 80.10 & 90.40 & 84.50 & 87.70 & 81.10\\
FusionReID \cite{wang2025unity} & 86.70 & 80.70 & 87.90 & 80.00 & 89.70 & 84.20 & 86.50 & 80.90\\
CLIP-ReID \cite{li2023clip} & 85.36 & 79.79 & 85.64 & 79.08 & 89.14 & 84.23 & 86.50 & 79.55 \\
PCL-CLIP \cite{li2023prototypical} & 79.80 & 72.20 & 81.12 & 72.40 & 87.14 & 77.70 & 84.19 & 73.89\\
V2E AGREIDv2 \cite{nguyen2024ag} & 88.77 & 80.72 & 87.86 & 78.51 & 93.62 & 84.85 & 88.61 & 80.11 \\
Explain \cite{nguyen2023ag} & 87.70 & 79.00 & 87.35 & 78.24 & \cellcolor{SecondBest}\underline{93.67} & 83.14 & 87.73 & 79.08 \\
VDT  \cite{zhang2024view} & 86.46 & 79.13 & 86.14 & 78.12 & 90.00 & 82.21 & 85.26 & 78.52 \\
SeCap \cite{wang2025secap}  & 88.12 & 80.84 & 88.24 & 79.99 & 91.44 & 84.01 & 87.56 & 80.15 \\
SD-ReID \cite{wang2025sdreid} & 87.04 & 80.61 & 86.74 & 79.24 & 90.86 & 84.06 & 87.86 & 81.01\\
LATex \cite{zhang2025latex} & \cellcolor{SecondBest}\underline{89.13} & \cellcolor{SecondBest}\underline{83.50} & \cellcolor{SecondBest}\underline{89.01} & \cellcolor{SecondBest}\underline{82.85} & 91.35 & \cellcolor{SecondBest}\underline{86.35} & \cellcolor{SecondBest}\underline{89.32} & \cellcolor{SecondBest}\underline{83.30}\\
\hline
\textbf{GeoReID} & \cellcolor{Best}\textbf{91.26} & \cellcolor{Best}\textbf{85.52} 
              & \cellcolor{Best} \textbf{90.34} & \cellcolor{Best} \textbf{83.88}
              & \cellcolor{Best} \textbf{93.21} & \cellcolor{Best} \textbf{88.03}
              & \cellcolor{Best} \textbf{90.95} & \cellcolor{Best} \textbf{84.79} \\
\hline
\end{tabular}

\end{table}

Results on AG-ReIDv2 are summarized in Table~\ref{Tbl:agreidv2_main}, which reports performance across four cross view protocols (A$\rightarrow$G, G$\rightarrow$A, A$\rightarrow$W, and W$\rightarrow$A).
GeoReID achieves the best overall performance across all protocols.
Under the most challenging A$\rightarrow$G setting, GeoReID reaches 91.26\% Rank-1 accuracy and 85.52\% mAP, outperforming all competing methods.
Consistent improvements are also observed under G$\rightarrow$A, where GeoReID achieves 90.34\% Rank-1 accuracy and 83.88\% mAP.
Notably, GeoReID yields strong gains under the A$\rightarrow$W and W$\rightarrow$A protocols, which involve low elevation wearable viewpoints and severe viewpoint asymmetry, achieving the highest mAP in both cases.
These results demonstrate that rectifying geometry induced distortion in the similarity space leads to more reliable cross view ranking under heterogeneous camera configurations.
\begin{table}[ht]
\centering
\scriptsize
\renewcommand\tabcolsep{2pt}
\renewcommand\arraystretch{1.12}

\caption{\textbf{Performance comparison on the CARGO dataset under metadata-free settings.}
The CARGO dataset does not provide camera geometry annotations.
We generate altitude, horizontal distance, and viewing angle using our vision-only meta predictor.
The same predicted geometry is provided to all geometry-aware baselines, while methods without geometry modeling are evaluated without geometry information.
Best results are highlighted as \protect\colorbox{Best}{\textbf{bold}}, and second-best results are highlighted as
\protect\colorbox{SecondBest}{\underline{underlined}}.}
\label{Tbl:cargo_main}
\begin{tabular}{l|cc|cc|cc|cc}
\hline
\multirow{2}{*}{\textbf{Model}}
 & \multicolumn{2}{c|}{\textbf{All}}
 & \multicolumn{2}{c|}{\textbf{G$\rightarrow$G}}
 & \multicolumn{2}{c|}{\textbf{A$\rightarrow$A}}
 & \multicolumn{2}{c}{\textbf{A$\rightarrow$G}} \\
\cline{2-9}
 & Rank-1 & mAP & Rank-1 & mAP & Rank-1 & mAP & Rank-1 & mAP \\
\hline
SBS \cite{he2023fastreid} & \scriptsize{50.32} & 43.09 & 72.31 & 62.99 & 67.50 & 49.73 & 31.25 & 29.00 \\
PCB \cite{sun2019learning} & 51.00 & 44.50 & 74.10 & 67.60 & 55.00 & 44.60 & 34.40 & 30.40 \\
BoT \cite{luo2019bag} & 54.81 & 46.49 & 77.68 & 66.47 & 65.00 & 49.79 & 36.25 & 32.56 \\
MGN \cite{wang2018learning} & 54.81 & 49.08 & 83.93 & 71.05 & 65.00 & 52.96 & 31.87 & 33.47 \\
VV \cite{kuma2019vehicle} & 45.83 & 38.84 & 72.31 & 62.99 & 67.50 & 49.73 & 31.25 & 29.00 \\
AGW \cite{ye2021deep} & 60.26 & 53.44 & 81.25 & 71.66 & 67.50 & 56.48 & 43.57 & 40.90 \\
BAU \cite{cho2024generalizable} & 45.20 & 38.40 & 61.60 & 51.20 & 50.00 & 42.60 & 40.40 & 36.70 \\
PAT \cite{ni2023part} & 37.90 & 15.30 & 52.70 & 24.20 & 50.00 & 23.10 & 35.10 & 15.50 \\
DTST \cite{wang2024dynamic} & 64.42 & 55.73 & 78.57 & 72.40 & 80.00 & 63.31 & 50.53 & 43.49 \\
ViT \cite{dosovitskiy2020image} & 61.54 & 53.54 & 82.14 & 71.34 & 80.00 & 64.47 & 43.13 & 40.11 \\
VDT \cite{zhang2024view} & 64.10 & 55.20 & 82.14 & 71.59 & 82.50 & 66.83 & 48.12 & 42.76 \\
TransReID \cite{he2021transreid} & 73.70 & \cellcolor{SecondBest}\underline{64.70} & 85.70 & 77.90 & \cellcolor{Best}\textbf{85.00} & \cellcolor{Best}\textbf{71.80} & 64.40 & 55.90 \\
FusionReID \cite{wang2025unity} & 67.90 & 61.50 & 85.70 & 79.40 & 80.00 & \cellcolor{SecondBest}\underline{69.30} & 48.30 & 53.10 \\
CLIP-ReID \cite{li2023clip} & 68.27 & 64.25 & 84.82 & \cellcolor{SecondBest}\underline{80.80} & 75.00 & 65.42 & 55.62 & 53.83 \\
PCL-CLIP \cite{li2023prototypical} & 67.31 & 60.93 & 84.82 & 76.00 & 70.00 & 60.75 & 54.43 & 51.43 \\
SD-ReID \cite{wang2025sdreid} & \cellcolor{SecondBest}\underline{70.48} & 61.42 & 81.25 & 74.08 & 82.50 & 67.70 & 48.75 & 46.37 \\
GSAlign \cite{li2025gsalign} & 65.06 & 57.95 & 83.04 & 73.86 & 80.00 & 65.55 & 64.89 & \cellcolor{SecondBest}\underline{61.55} \\
SeCap \cite{wang2025secap} & 68.59 & 60.19 & \cellcolor{SecondBest}\underline{86.61} & 75.42 & 80.00 & 68.08 & \cellcolor{SecondBest}\underline{69.43} & 58.94 \\
\hline
\textbf{GeoReID}
 & \cellcolor{Best}\textbf{71.79} & \cellcolor{Best}\textbf{63.69}
 & \cellcolor{Best}\textbf{88.04} & \cellcolor{Best}\textbf{79.14}
 & \cellcolor{SecondBest}\underline{82.51} & 68.20
 & \cellcolor{Best}\textbf{72.02} & \cellcolor{Best}\textbf{64.61} \\
\hline
\end{tabular}
\end{table}

\par
\begin{table}[ht]
\centering
\scriptsize
\renewcommand\tabcolsep{4pt}
\renewcommand\arraystretch{1.1}

\caption{\textbf{Performance comparison on the DETReIDX dataset across three evaluation protocols.}
Results are reported using the best validation checkpoint.
Best values per column are shown in \textbf{bold}.}
\label{Tbl:detreidx_main}

\begin{tabular}{l|cc|cc|cc}
\hline
\multirow{2}{*}{\textbf{Method}}
 & \multicolumn{2}{c|}{\textbf{G$\rightarrow$A}}
 & \multicolumn{2}{c|}{\textbf{A$\rightarrow$G}}
 & \multicolumn{2}{c}{\textbf{A$\rightarrow$A}} \\
\cline{2-7}
 & Rank-1 & mAP & Rank-1 & mAP & Rank-1 & mAP \\
\hline
SBS \cite{he2023fastreid} & 46.74 & 13.53 & 15.59 & 15.51 & 13.74 & 9.09 \\
MGN \cite{wang2018learning} & 46.48 & 14.62 & 15.10 & 16.33 & 13.06 & 8.97 \\
AGW \cite{ye2021deep} & 30.59 & 8.56 & 12.30 & 13.13 & 10.85 & 6.84 \\
VDT \cite{zhang2024view} & \cellcolor{SecondBest}\underline{50.36} & \cellcolor{SecondBest}\underline{16.37} & \cellcolor{SecondBest}\underline{18.14} & \cellcolor{SecondBest}\underline{18.55} & \cellcolor{SecondBest}\underline{15.27} & \cellcolor{SecondBest}\underline{9.35} \\
SeCap \cite{wang2025secap} & \cellcolor{Best}\textbf{51.63} & 14.86 & 17.05 & 17.77 & 14.03 & 9.12 \\
\hline
\textbf{GeoReID}
 & 48.87 & \cellcolor{Best}\textbf{21.28}
 & \cellcolor{Best}\textbf{20.84} & \cellcolor{Best}\textbf{23.13}
 & \cellcolor{Best}\textbf{16.28} & \cellcolor{Best}\textbf{12.97} \\
\hline
\end{tabular}
\end{table}
\par

Quantitative results on the CARGO dataset are reported in Table~\ref{Tbl:cargo_main}.
CARGO represents a realistic metadata free aerial-ground ReID scenario, where camera geometry annotations are unavailable and must be estimated.
GeoReID achieves the best overall performance under the All protocol, with 71.79\% Rank-1 accuracy and 63.69\% mAP.
Under the most challenging A$\rightarrow$G protocol, GeoReID attains 72.02\% Rank-1 accuracy and 64.61\% mAP, outperforming all competing methods.
This setting involves the largest viewpoint, scale, and perspective discrepancies, highlighting the effectiveness of geometry conditioned similarity rectification for difficult cross-view matching.
Performance on same view protocols (G$\rightarrow$G and A$\rightarrow$A) remains competitive, indicating that geometry conditioning does not compromise same view discrimination.
These results show that GeoReID remains effective even when camera geometry is predicted rather than provided.
Results on the DETReIDX dataset are summarized in Table~\ref{Tbl:detreidx_main}.
DETReIDX is particularly challenging due to noisy detections, low image resolution, and extreme aerial-ground viewpoint differences, leading to low absolute performance across all methods.
GeoReID achieves the best performance under the A$\rightarrow$G protocol, obtaining 20.84\% Rank-1 accuracy and 23.13\% mAP.
Across all three evaluation protocols, GeoReID consistently achieves the highest mAP, indicating more reliable ranking quality under severe geometric distortion.
Under the G$\rightarrow$A setting, GeoReID attains the highest mAP despite slightly lower Rank-1 accuracy, while performance on the A$\rightarrow$A protocol also improves consistently.
These results confirm that explicitly rectifying geometry induced similarity distortion is particularly beneficial in extremely challenging cross view scenarios.

\subsection{Ablation Studies and Analysis}

\textbf{Component Ablation Analysis}

\begin{table}[t]
\centering
\scriptsize
\caption{Ablation and geometry embedding analysis on DETReIDX}
\label{tab:detreidx}
\resizebox{\columnwidth}{!}{%
\begin{tabular}{l|cc|cc|cc}
\hline
\multirow{2}{*}{Method}
 & \multicolumn{2}{c|}{G$\rightarrow$A}
 & \multicolumn{2}{c|}{A$\rightarrow$G}
 & \multicolumn{2}{c}{A$\rightarrow$A} \\
\cline{2-7}
 & R1 & mAP & R1 & mAP & R1 & mAP \\
\hline
Baseline (SeCap) & 51.63 & 14.86 & 17.05 & 17.77 & 14.03 & 9.12  \\
+ GCPG                & 46.33 & 19.47 & 19.13 & 21.30 & 13.73 & 11.74 \\
+ GIQT              & 46.59 & 19.82 & 19.63 & 22.04 & 15.30 & 12.14 \\
+ GCPG + GIQT         & 47.42 & 20.32 & 20.06 & 22.76 & 15.85 & 12.09 \\
\hline
\multicolumn{7}{l}{\textit{Geometry Embedding Analysis}} \\
\hline
w/o Cam             & 46.57 & 19.38 & 18.46 & 21.04 & 14.19 & 11.93 \\
w/o Alt             & 47.51 & 20.36 & 20.11 & 22.56 & 14.81 & 11.77 \\
w/o Ang             & 47.31 & 20.15 & 20.22 & 22.54 & 14.94 & 12.06 \\
\hline
\end{tabular}}
\end{table}

\begin{table}[t]
\centering
\scriptsize
\caption{Ablation study on AG-ReID}
\label{tab:agreid}

{%
\begin{tabular}{l|cc|cc}
\hline
\multirow{2}{*}{Method}
 & \multicolumn{2}{c|}{A$\leftrightarrow$G}
 & \multicolumn{2}{c}{G$\leftrightarrow$A} \\
\cline{2-5}
 & R1 & mAP & R1 & mAP \\
\hline
Baseline (SeCap)    & 84.03 & 76.20 & 87.21 & 79.19 \\
+ GCPG                & 86.46 & 78.68 & 89.50 & 82.24 \\
+ GIQT              & 85.47 & 77.36 & 87.97 & 79.73 \\
+ GCPG + GIQT         & 87.02 & 79.46 & 90.64 & 83.55 \\
\hline
\multicolumn{5}{l}{\textit{Geometry Embedding Analysis}} \\
\hline
w/o Cam             & 86.37 & 79.39 & 89.50 & 82.61 \\
w/o Alt             & 84.50 & 75.99 & 86.38 & 78.77 \\
w/o Ang             & 86.37 & 78.83 & 88.88 & 81.52 \\
\hline
\end{tabular}}
\end{table}

\begin{table}[t]
\centering
\caption{Ablation study on AG-ReIDv2}
\label{tab:agreidv2}
\fontsize{8.5pt}{10pt}\selectfont
\resizebox{\columnwidth}{!}{%
\begin{tabular}{l|cc|cc|cc|cc}
\hline
\multirow{2}{*}{Method}
 & \multicolumn{2}{c|}{A$\leftrightarrow$G}
 & \multicolumn{2}{c|}{G$\leftrightarrow$A}
 & \multicolumn{2}{c|}{A$\leftrightarrow$W}
 & \multicolumn{2}{c}{W$\leftrightarrow$A} \\
\cline{2-9}
 & R1 & mAP & R1 & mAP & R1 & mAP & R1 & mAP \\
\hline
Baseline (SeCap)    & 88.12 & 80.84 & 88.24 & 79.99 & 91.44 & 84.01 & 87.56 & 80.15 \\
+ GCPG                & 90.24 & 84.41 & 88.90 & 83.45 & 93.30 & 87.65 & 90.51 & 84.21 \\
+ GIQT              & 88.76 & 82.59 & 88.13 & 81.70 & 91.13 & 85.63 & 88.76 & 82.59 \\
+ GCPG + GIQT         & 90.15 & 84.49 & 89.45 & 83.18 & 92.80 & 87.54 & 90.77 & 84.71 \\
\hline
\multicolumn{9}{l}{\textit{Geometry Embedding Analysis}} \\
\hline
w/o Cam             & 90.32 & 84.71 & 89.84 & 83.54 & 93.30 & 87.80 & 90.94 & 84.55 \\
w/o Alt             & 88.67 & 82.61 & 88.57 & 81.76 & 91.58 & 85.71 & 88.85 & 82.58 \\
w/o Ang             & 90.15 & 84.38 & 89.84 & 83.65 & 92.76 & 87.83 & 90.43 & 84.86 \\
\hline
\end{tabular}}
\end{table}

\begin{table}[t]
\centering
\caption{Ablation study on CARGO}
\label{tab:cargo}
\fontsize{8.5pt}{10pt}\selectfont
\resizebox{\columnwidth}{!}{%
\begin{tabular}{l|cc|cc|cc|cc}
\hline
\multirow{2}{*}{Method}
 & \multicolumn{2}{c|}{All}
 & \multicolumn{2}{c|}{G$\rightarrow$G}
 & \multicolumn{2}{c|}{A$\rightarrow$A}
 & \multicolumn{2}{c}{A$\rightarrow$G} \\
\cline{2-9}
 & R1 & mAP & R1 & mAP & R1 & mAP & R1 & mAP \\
\hline
Baseline (SeCap) & 68.59 & 60.19 & 86.61 & 75.42 & 80.00 & 68.08 & 69.43 & 58.94\\
+ GCPG & 70.83 & 63.94 & 86.61 & 80.79 & 85.00 & 69.08 & 71.28 & 68.41 \\
+ GIQT   & 72.76 & 64.62 & 88.39 & 79.40 & 82.50 & 68.90 & 72.34 & 68.52 \\
\hline
\multicolumn{9}{l}{\textit{Geometry Embedding Analysis}} \\
\hline
w/o Cam   & 69.87 & 63.29 & 85.71 & 80.31 & 82.50 & 69.55 & 67.02 & 65.23 \\
w/o Alt     & 69.23 & 63.60 & 84.82 & 80.36 & 85.00 & 68.50 & 69.15 & 68.95 \\
w/o Ang     & 69.87 & 63.97 & 86.61 & 81.35 & 80.00 & 66.74 & 69.15 & 67.73 \\
\hline
\end{tabular}}
\end{table}

\label{sec:ablation}

This section analyzes the contribution of each proposed component and geometry cue across four benchmarks with progressively increasing geometric difficulty: AG-ReID, AG-ReIDv2, CARGO, and DETReIDX.

\subsubsection{Component Level Analysis}

We first evaluate the individual and combined effects of geometry conditioned prompts and the GIQT.
As shown in Table~\ref{tab:agreid}, adding GCPG improves Rank-1 accuracy from $84.03\%$ to $86.46\%$, confirming that global representation priors benefit from explicit geometry conditioning.
GIQT alone provides smaller but consistent gains, indicating the presence of local similarity distortion even when global features are unchanged.
Combining GCPG and GIQT yields the best results across both A$\leftrightarrow$G and G$\leftrightarrow$A protocols, demonstrating complementary effects between global adaptation and local similarity rectification.
Table~\ref{tab:agreidv2} shows that the same trend holds under more diverse camera configurations, including wearable views.
GCPG provides strong improvements across all protocols, while GIQT yields consistent gains, particularly in cross view settings.
The combined GCPG+GIQT model achieves the best overall performance, with notable improvements under the challenging A$\leftrightarrow$G and W$\leftrightarrow$A protocols, where viewpoint asymmetry is most severe.
CARGO represents a metadata free scenario where geometry must be predicted.
As shown in Table~\ref{tab:cargo}, GIQT yields larger gains than GCPG under the most challenging A$\rightarrow$G protocol, improving Rank-1 accuracy from $71.28\%$ to $72.34\%$.
This suggests that under extreme cross view mismatch and imperfect geometry estimation, correcting similarity space distortion becomes more critical than global prompt adaptation alone.
DETReIDX is characterised by noisy detections, low resolution, and extreme aerial-ground viewpoint differences.
As shown in Table~\ref{tab:detreidx}, GCPG and GIQT individually provide modest gains, while their combination achieves the highest mAP across all protocols.
Although absolute performance remains low due to dataset difficulty, the consistent improvement in mAP indicates more reliable ranking under severe geometric distortion.

\textbf{Geometry Embedding Analysis}

We further analyze (Table~\ref{tab:detreidx}, \ref{tab:agreid}, \ref{tab:agreidv2}, \ref{tab:cargo}) the contribution of individual geometry cues by removing camera identity, altitude, and viewing-angle encoding.
Across all datasets, removing altitude or viewing angle information leads to larger performance degradation than removing camera identity.
For example, on AG-ReID (Table~\ref{tab:agreid}), removing altitude reduces A$\leftrightarrow$G Rank-1 accuracy from $87.02\%$ to $84.50\%$, while removing camera identity results in a smaller drop.
Similar trends are observed on AG-ReIDv2, CARGO, and DETReIDX.
These results indicate that geometry cues altitude and viewing angle are the dominant factors governing geometry induced similarity distortion, while camera identity primarily provides coarse contextual bias.


\subsection{Analysis of Design Choices}
\label{sec:design_analysis}

This subsection provides empirical justification for the architectural and modeling choices of the proposed framework.
All analyses are conducted consistently across AG-ReID, AG-ReIDv2, CARGO, and DETReIDX, covering both metadata rich and metadata free settings, as well as moderate to extreme viewpoint disparity.

\textbf{Structure of Geometry Induced Distortion}
\begin{figure}
    \centering
    \includegraphics[width=\linewidth]{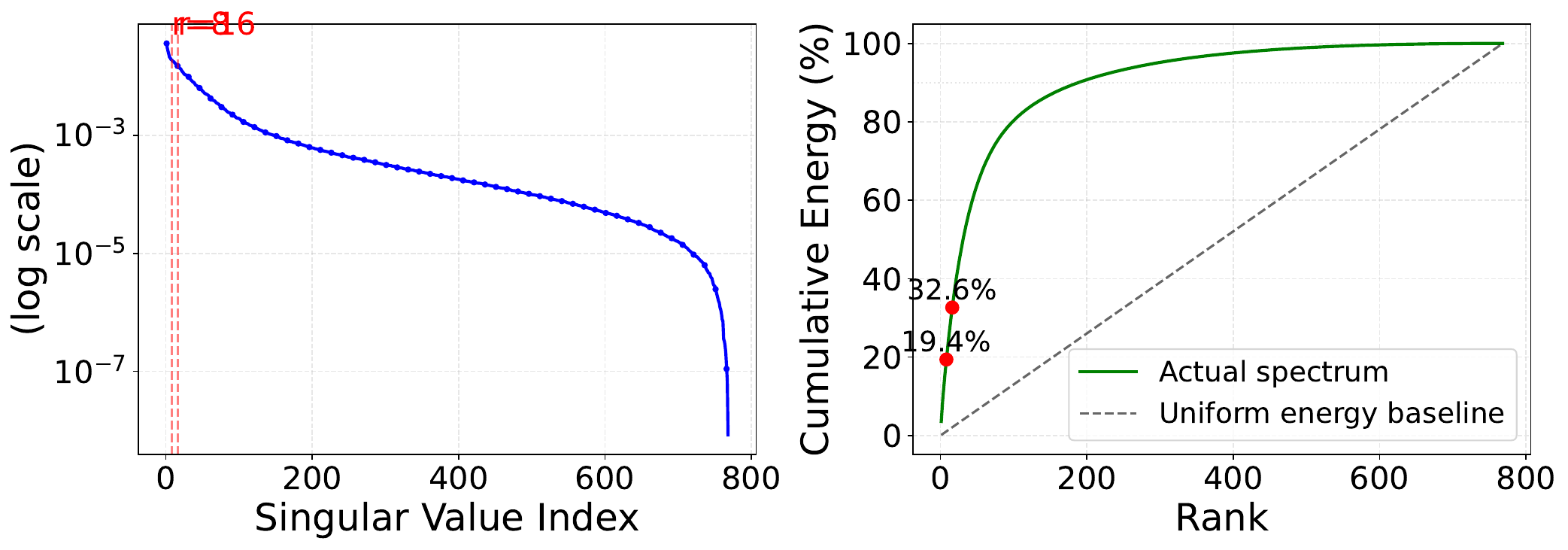}
    \caption{Singular value spectrum and cumulative energy distribution of cross view feature covariance difference, showing highly anisotropic geometry induced distortion.}
    \label{fig:spectrum}
\end{figure}

We first analyze the structure of cross view geometric distortion by examining the singular value spectrum of the covariance difference between aerial and ground features.
As shown in Fig.~\ref{fig:spectrum}, the spectrum exhibits pronounced decay, indicating that geometry induced distortion is highly anisotropic rather than uniformly distributed across feature dimensions.
Notably, the top-$8$ singular components capture approximately $19.4\%$ of the total energy, while the top-$16$ components capture $32.6\%$, far exceeding a uniform-energy baseline.
This observation motivates the low rank residual formulation adopted in the GIQT, which targets dominant geometry sensitive directions without overfitting higher order variations.

\textbf{Sensitivity to GIQT Rank}

\begin{figure}
    \centering
    \includegraphics[width=\linewidth]{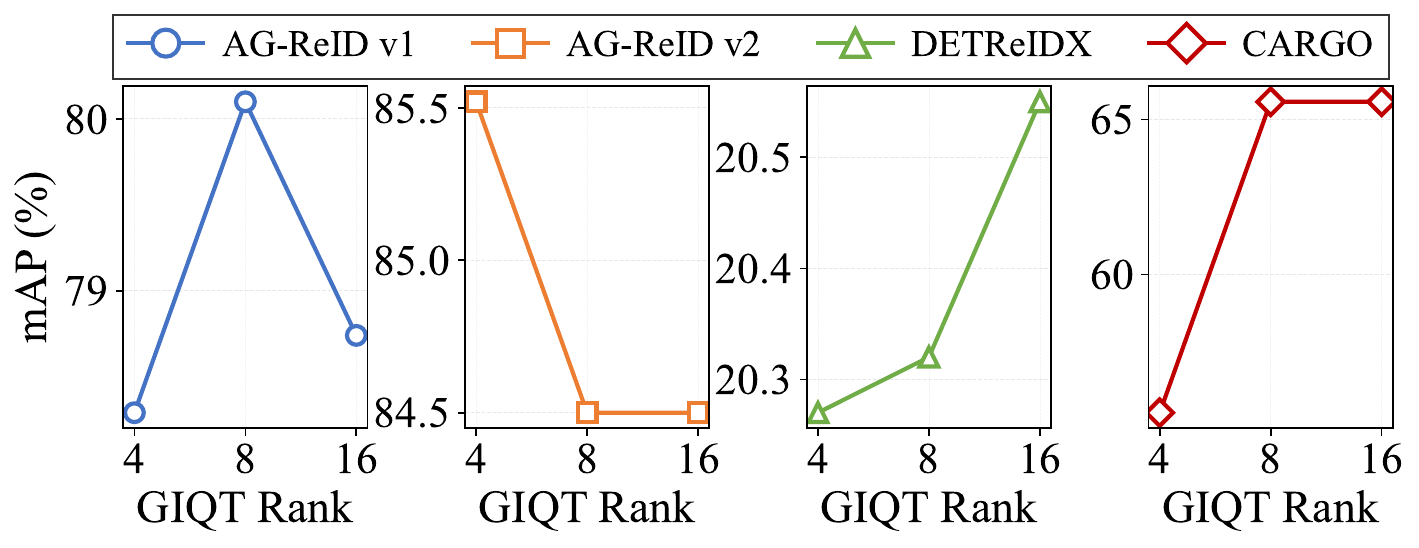}
    \caption{Performance sensitivity with respect to GIQT rank across datasets. Low rank correction (8--16) yields optimal performance.}
    \label{fig:giqt_rank}
\end{figure}
Figure~\ref{fig:giqt_rank} analyzes the effect of the GIQT rank on performance across AG-ReIDv1, AG-ReIDv2, DETReIDX, and CARGO.
Across all datasets, performance improves when increasing the rank from 4 to 8 and either saturates or slightly degrades beyond rank 16.
In particular, AG-ReIDv1 achieves its peak mAP at rank 8, while AG-ReIDv2 and CARGO exhibit marginal gains up to rank 16.
DETReIDX, which features more severe geometric distortion, also benefits most from low-to-moderate ranks.
These results indicate that geometry induced distortion is dominated by a small number of directions and that low-rank correction (8--16) is sufficient, while higher ranks provide diminishing returns and risk of overfitting.

\textbf{Sensitivity to geometry conditioned prompt length}

\begin{figure}
    \centering
    \includegraphics[width=\linewidth]{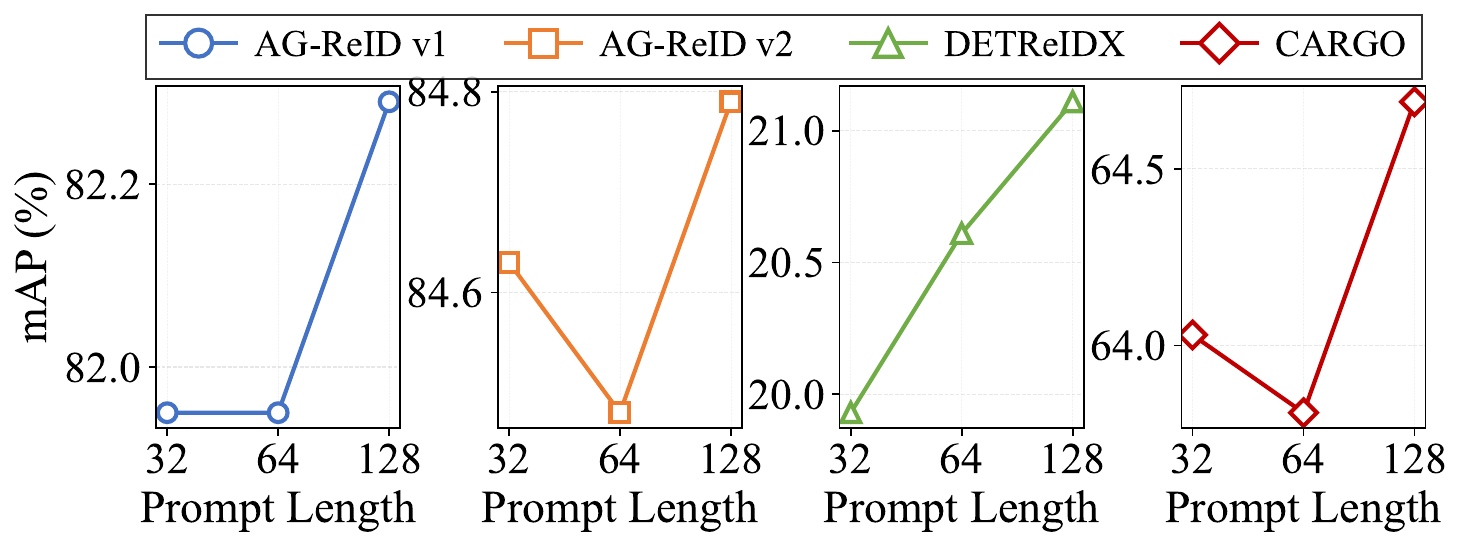}
    \caption{Effect of geometry conditioned prompt length across datasets. Performance remains stable over a wide range of prompt sizes.}
    \label{fig:prompt_len}
\end{figure}
Figure~\ref{fig:prompt_len} evaluates the impact of geometry conditioned prompt length.
Performance remains stable across a wide range of prompt sizes (32--128 tokens) on all datasets.
AG-ReIDv1 and AG-ReIDv2 show minor fluctuations, while DETReIDX and CARGO exhibit slightly better performance with longer prompts.
These results demonstrate that the proposed GCPG mechanism is robust to prompt length and does not require careful tuning, indicating that geometry information is encoded compactly rather than relying on excessive prompt capacity.

\textbf{Hidden Dimension Sensitivity}
\begin{figure}
    \centering
    \includegraphics[width=\linewidth]{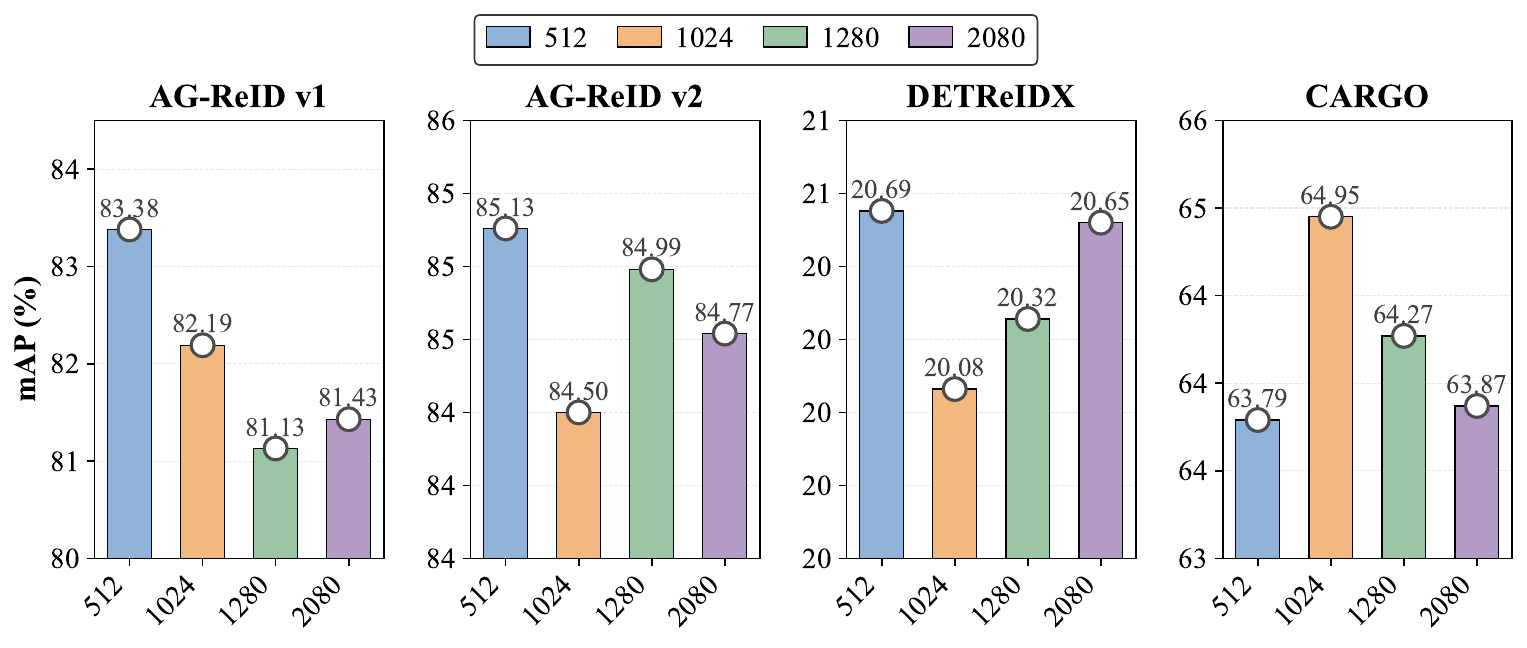}
    \caption{Effect of hidden dimension size across datasets. Increasing feature dimensionality beyond 512 does not yield consistent gains, indicating that geometry induced similarity distortion—not model capacity is the dominant bottleneck.}
\label{fig:hidden_dim}
\end{figure}
Figure~\ref{fig:hidden_dim} analyzes the effect of increasing the hidden dimension across all four benchmarks.
Across AG-ReID, AG-ReIDv2, CARGO, and DETReIDX, increasing the hidden dimension beyond 512 does not yield consistent improvements and in several cases slightly degrades performance.
This behavior indicates that cross view performance degradation under extreme aerial-ground geometry is not caused by insufficient representational capacity.
Simply increasing feature dimensionality does not resolve the dominant failure mode.
Instead, the limiting factor lies in the mismatch of the similarity space induced by geometric distortion.
This observation provides further justification for explicitly rectifying similarity computation via geometr conditioned query-key transformation, rather than increasing model capacity or relying on wider embeddings.

\textbf{Sensitivity to Prompt Scaling}
\begin{figure}
    \centering
    \includegraphics[width=\linewidth]{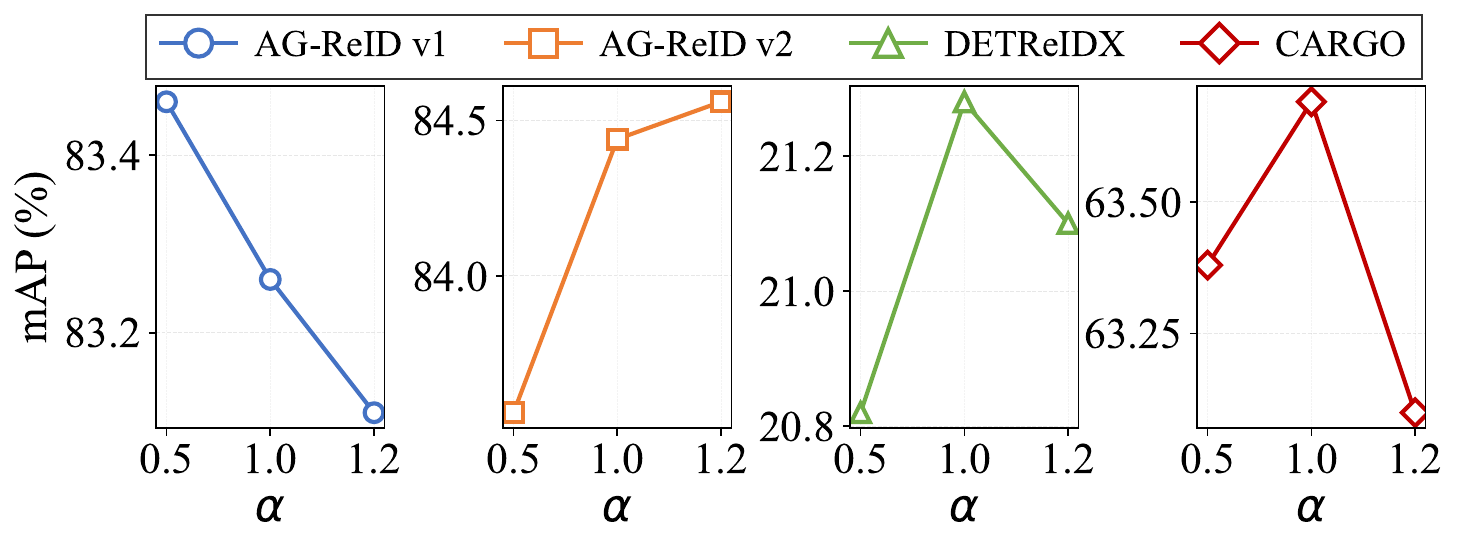}
    \caption{Sensitivity to geometry-conditioned prompt scaling factor $\alpha$ across datasets. Performance peaks near $\alpha=1.0$, indicating that geometry prompts should bias representations rather than dominate identity features.}
    \label{fig:GCPG_alpha}
\end{figure}
Figure~\ref{fig:GCPG_alpha} analyses the effect of the geometry-conditioned prompt scaling factor $\alpha$ across all four benchmarks. Across datasets, performance consistently peaks around $\alpha = 1.0$ and degrades when the prompt contribution is either underscaled or over-emphasised.
When $\alpha$ is too small, geometry-conditioned prompts fail to provide sufficient bias to compensate for viewpoint-induced distortion. Conversely, when $\alpha$ is too large, geometry cues begin to dominate identity-discriminative features, leading to reduced retrieval performance. This behaviour is particularly evident on CARGO and DETReIDX, where extreme viewpoint disparity and noisy geometry estimation amplify the negative effects of over-scaling.
These results confirm that geometry-conditioned prompts should act as a structured bias rather than a dominant representation component. The residual prompt formulation enables controlled adaptation, allowing geometry information to guide feature extraction without overwhelming identity semantics.


\vspace{2pt}
\noindent\textbf{Corruption definitions.}

\begin{table}[t]
\centering
\scriptsize
\renewcommand\arraystretch{1.2}

\caption{\textbf{Robustness to discretized geometry metadata corruption across datasets.}
Results are reported under Protocol~1 (A$\leftrightarrow$G) for AG-ReID and A$\rightarrow$G for DetReIDX.
AG-ReID uses 3 altitude bins and 3 viewing-angle bins, while DetReIDX uses 6 altitude bins and 3 viewing-angle bins.
All corruptions are applied \emph{only at inference time} to geometry inputs, using models trained with clean geometry.
$\Delta$R1 (Baseline) is computed relative to the geometry-agnostic SeCap baseline
(84.03\% Rank-1 for AG-ReID and 17.05\% Rank-1 for DetReIDX).}
\label{tab:robustness_combined}

\begin{tabular}{l|cc|cc}
\hline
\multirow{2}{*}{\textbf{Corruption Type}}
& \multicolumn{2}{c|}{\textbf{AG-ReID}}
& \multicolumn{2}{c}{\textbf{DetReIDX}} \\
\cline{2-5}
& Rank-1 & $\Delta$R1 (Base)
& Rank-1 & $\Delta$R1 (Base) \\
\hline

Clean geometry
& 87.02 & +2.99
& 20.84 & +3.79 \\

Random bin flip (Altitude)
& 86.12 & +2.09
& 20.12 & +3.07 \\

Random bin flip (Angle)
& 85.71 & +1.68
& 19.63 & +2.58 \\

Joint bin perturbation
& 84.98 & +0.95
& 19.05 & +2.00 \\

Biased altitude shift
& 84.61 & +0.58
& 18.67 & +1.62 \\

Stale geometry
& 85.23 & +1.20
& 19.34 & +2.29 \\

Wrong geometry
& 84.17 & +0.14
& 18.21 & +1.16 \\
\hline
\end{tabular}
\end{table}
\emph{Random bin flip (Altitude / Angle)} randomly shifts the corresponding geometry bin by $\pm1$, clipped to valid bin boundaries.
\emph{Joint bin perturbation} applies independent $\pm1$ shifts to both altitude and angle bins.
\emph{Biased altitude shift} applies a fixed $+1$ altitude-bin offset to all samples.
\emph{Stale geometry} assigns geometry bins from a temporally adjacent or nearby sample, simulating delayed or asynchronous metadata.
\emph{Wrong geometry} randomly shuffles geometry bins across identities, breaking geometry–image correspondence while preserving the marginal bin distribution.
All perturbations are applied independently per image and consistently to both query and gallery samples.

\begin{figure}
    \centering
    \includegraphics[width=\linewidth]{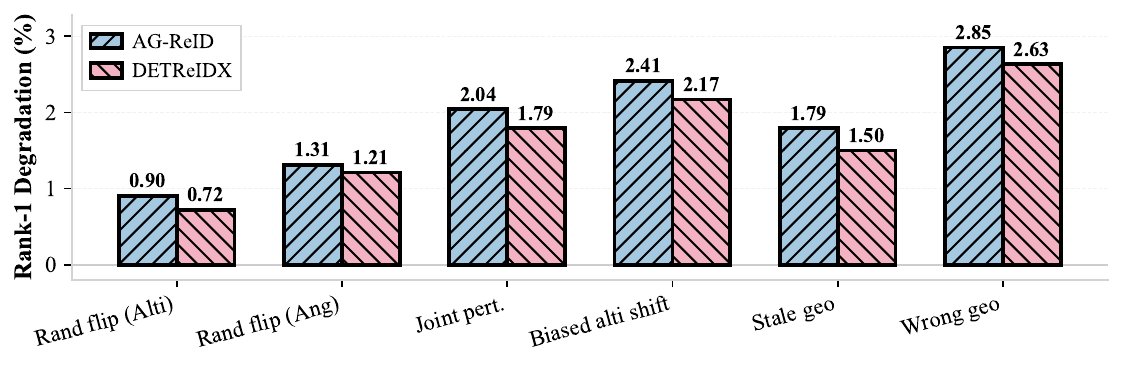}
    \caption{Robustness to discretized geometry metadata corruption.
    The y-axis shows Rank-1 degradation relative to clean geometry.
    All corruptions are applied only at inference time using models trained with clean geometry.
    Results are reported for AG-ReID (A$\leftrightarrow$G) and DetReIDX (A$\rightarrow$G).}
    \label{fig:geometry_corruption_robustness}
\end{figure}

\subsection{Visualization Analysis}
\textbf{Qualitative Retrieval Analysis}

\begin{figure}
    \centering
    \includegraphics[width=\linewidth]{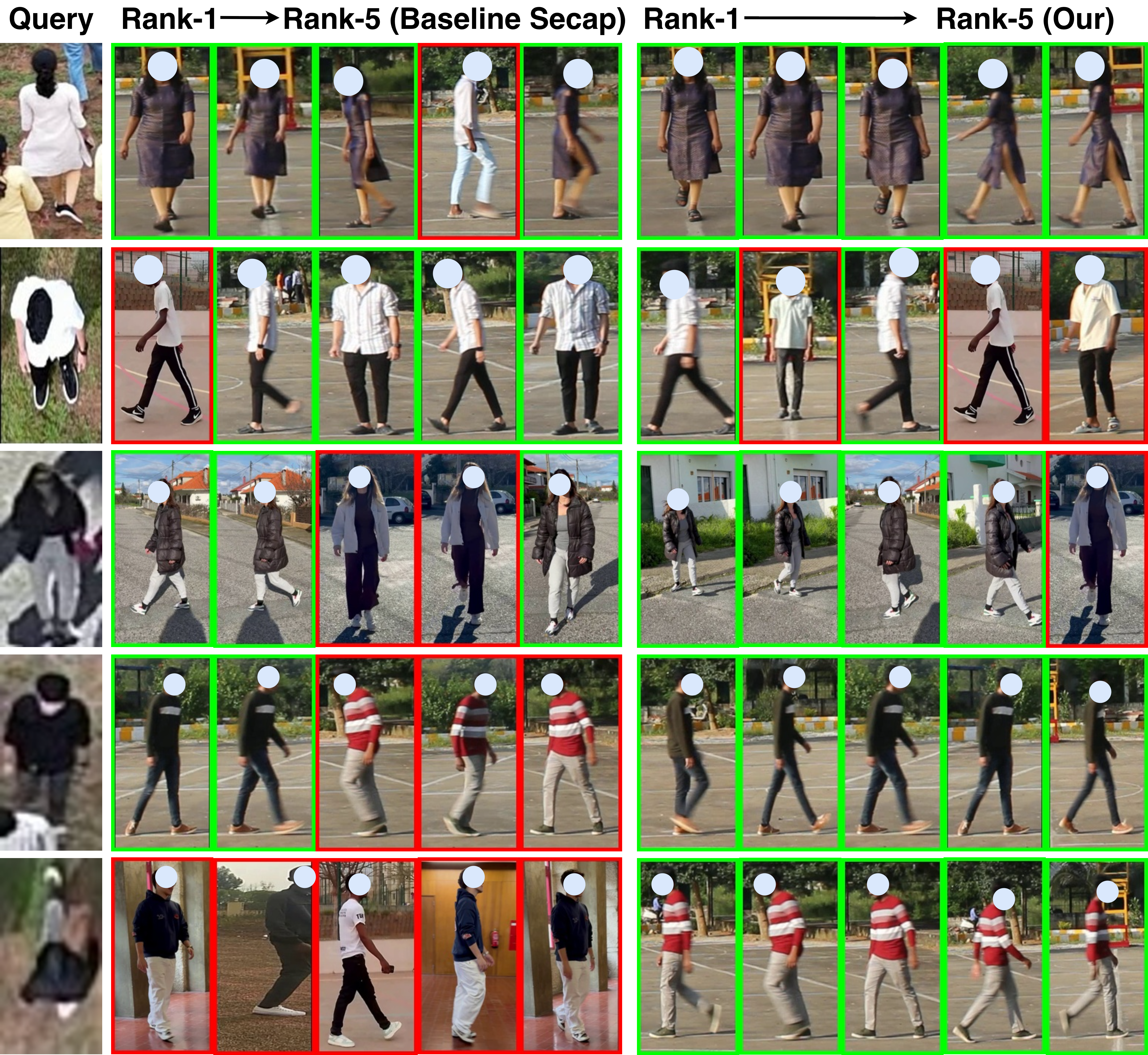}
    \caption{Qualitative comparison of Rank-1 to Rank-5 retrieval results between the baseline (SeCap) and the proposed method. The baseline suffers from incorrect high-rank matches due to geometry-induced similarity distortion, while the proposed method produces more consistent and correct ranking under extreme aerial--ground viewpoint differences.}
    \label{fig:qualitative}
\end{figure}

Figure~\ref{fig:qualitative} presents representative Rank-1 to Rank-5 retrieval results comparing the baseline (SeCap) and the proposed method under challenging aerial--ground matching scenarios. Correct matches are highlighted in green, while incorrect matches are shown in red.
The baseline frequently retrieves visually similar but incorrect identities under extreme viewpoint changes, particularly when aerial queries induce strong foreshortening, scale compression, or body-part displacement. In these cases, similarity scores are dominated by spurious local alignments, leading to incorrect high-rank matches despite overall appearance resemblance.
In contrast, the proposed method consistently retrieves correct identities at higher ranks. Notably, improvements are observed not only at Rank-1 but also across the entire top-$k$ list, indicating more reliable similarity ordering rather than isolated corrections. This behavior suggests that geometry-induced distortion in the similarity space is explicitly mitigated, enabling more consistent alignment of corresponding body regions across views.
These qualitative results complement the quantitative gains, illustrating that geometry-conditioned similarity rectification improves ranking stability under extreme cross-view viewpoint discrepancies.

\begin{figure}
    \centering
    \includegraphics[width=\linewidth]{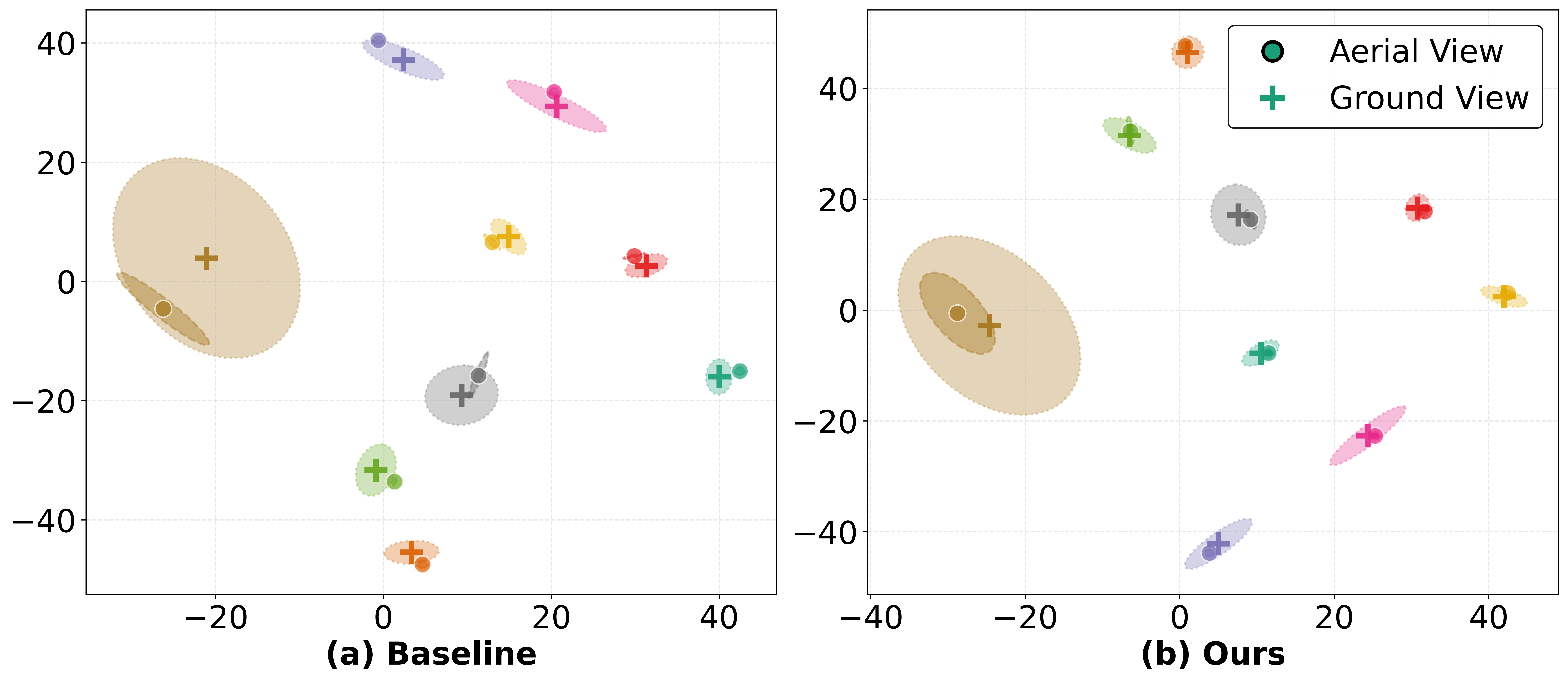}
    \caption{2D t-SNE visualization of aerial and ground view feature embeddings for the baseline and our proposed method. Circles ($\bullet$) represent the geometric mean aerial-view samples, and plus symbols ($+$) represent ground-view samples, while the shaded areas provide the ellipses that contain 90\% of the corresponding data points. Colors indicate identity labels.}
    \label{fig:tsne}
\end{figure}

\subsection{Feature Space Visualization}

Figure~\ref{fig:tsne} provides the learned feature distributions of the baseline model and the proposed method using t-SNE. Each color denotes a  identity, while circles and crosses represent aerial and ground views, respectively.
In the baseline model, features corresponding to the same identity but different views exhibit noticeable separation and overlap with other identities. This indicates that although identity-level clustering exists, cross-view alignment remains inconsistent, leading to fragmented clusters and view-dependent feature drift.
In contrast, the proposed method produces more compact and well-aligned clusters across views. Features from aerial and ground cameras corresponding to the same identity are more tightly grouped, while inter-identity separation is preserved. Importantly, this improvement is not achieved by collapsing representations, but by reducing cross-view dispersion within each identity cluster.
While t-SNE is a qualitative visualization tool, the observed reduction in cross-view overlap and improved cluster compactness are consistent with the quantitative gains reported earlier. These results suggest that geometry-conditioned prompt adaptation and similarity-space rectification jointly encourage more view-consistent representations.

\section{Conclusion}

This work identified the \emph{geometry-induced distortion in the similarity space} factor as a key failure mode in aerial--ground person re-identification. As a consequence, we concluded that the geometry-invariant similarity assumption underlying attention does not hold, even when feature representations are partially aligned.
To address this issue, we introduced a geometry-conditioned similarity alignment framework that explicitly incorporates camera geometry into both global representation adaptation and local similarity computation. The proposed Geometry-Induced Query--Key Transformation (GIQT) rectifies dominant, anisotropic similarity distortions through a lightweight low-rank formulation, while geometry-conditioned prompts provide complementary global adaptation.
Extensive experiments across four benchmarks suggest that explicitly correcting the similarity space consistently improves robustness under extreme, corrupted, and unseen geometric conditions, with minimal computational overhead. These findings highlight the importance of geometry-aware similarity modeling for reliable cross-view recognition in \emph{real-world} aerial--ground scenarios.

\section*{Acknowledgements}
This work is funded by national funds through FCT – Fundação para a Ciência e a Tecnologia, I.P., and, when eligible, co-funded by EU funds under project/support UID/50008/2025 – Instituto de Telecomunicações, with DOI identifier \url{https://doi.org/10.54499/UID/50008/2025}.


\vfill

\end{document}